\documentclass{article} 
\usepackage[american]{babel}
\usepackage{authblk}
\usepackage{fullpage}
\usepackage{enumitem}
\usepackage{natbib} 
\usepackage{mathtools} 
\usepackage{booktabs} 
\usepackage{tikz} 

\usepackage{microtype}
\usepackage{graphicx}
\usepackage{caption}
\usepackage{subcaption}
\usepackage{booktabs} 

\usepackage{amsmath}
\usepackage{amssymb}
\usepackage{mathtools}
\usepackage{amsthm}
\usepackage{bbm}

\usepackage{hyperref}
\usepackage[capitalize,noabbrev]{cleveref}

\usepackage{algorithm}
\usepackage{algorithmic}

\usepackage{stackengine}
\usepackage{Definitions}
\usepackage{hyperref}
\usepackage[capitalize,noabbrev]{cleveref}

\newcommand{\comment}[1]{}

\newcommand{\given}{\,\mid\,}
\newcommand{\X}{\mathcal{X}}
\newcommand{\Y}{\mathcal{Y}}
\newcommand{\U}{\mathcal{U}}

\newcommand{\M}{\mathcal{M}}
\newcommand{\Hu}{\mathcal{H}}
\newcommand{\vecX}{\mathbf{X}}
\newcommand{\vecU}{\mathbf{U}}
\newcommand{\vecF}{\mathbf{F}}
\newcommand{\vecY}{\mathbf{Y}}
\newcommand{\vecy}{\mathbf{y}}

\newcommand{\Loss}{\mathcal{L}}

\newcommand{\xhdr}[1]{\vspace{3mm}\noindent {\bf #1.}}

\title{Counterfactual Inference of Second Opinions}
\author[1,2]{Nina~L.~Corvelo~Benz}
\author[1]{Manuel~Gomez~Rodriguez}
\affil[1]{%
	Max Planck Institute for Software Systems, \{ninacobe, manuelgr\}@mpi-sws.org
}
\affil[2]{%
	Department of Biosystems Science and Engineering, ETH Zurich
}
\date{}
\begin{document}
\maketitle

\begin{abstract}
%
%
Automated decision support systems that are able to infer second opi\-nions from experts can potentially facilitate a more efficient 
allocation of resources---they can help decide when and from whom to seek a se\-cond opi\-nion.
In this paper, we look at the design of this type of support systems from the perspective of counterfactual inference.
%
%
%
We focus on a multiclass classification setting and first show that, if experts make predictions on their own, the underlying causal 
mechanism generating their predictions needs to satisfy a desirable set invariant pro\-per\-ty.
Further, we show that, for any causal mechanism satisfying this property, there exists\- an equi\-va\-lent mechanism where the predictions
by each expert are generated by independent sub-mechanisms go\-ver\-ned by a common noise.
%
%
This motivates the design of a set invariant Gumbel-Max structural causal model where the structure of the noise governing the sub-mechanisms 
underpinning the model depends on an intuitive notion of si\-mi\-la\-ri\-ty between experts which can be estimated from data.
%
%
Experiments on both synthetic and real data show that our model can be used to infer se\-cond opinions more accurately than its non-causal counterpart.
\end{abstract}

\vspace{-2mm}
\section{Introduction} 
\label{sec:introduction}
\vspace{-2mm}

%

%
%
In decision making under uncertainty, seeking opinions from multiple human experts tends to improve the overall quality of the decisions.
For example, in medicine, second opinions have been shown valuable for establishing diagnoses and initiating treatment~\citep{burger2020outcomes} as 
well as reducing the number of unnecessary procedures~\citep{leape1989unnecessary,althabe2004mandatory}.
%
%
%
In machine learning, ground truth labels are determined by carefully aggregating multiple noisy labels provided by different experts~\citep{zhang2016learning} 
and inconsistencies between these noisy labels help developing more robust models~\citep{peterson2019human}.
Unfortunately, the timeliness and quality of the decisions is often compromised due to a shortage of experts, which prevents each decision to 
be informed by multiple experts'{} opinions.

In this context, we argue that the development of automated decision support systems that, given an expert'{}s opinion on a decision instance and a set of features, 
are able to infer other experts'{} opinions will enable a more efficient allocation of resources.
%
%
On the one hand, these sys\-tems could prevent (prioritize) seeking other experts'{} opinions when they are unlikely (likely) to bring new 
perspectives.
On the other hand, these systems could also help identify those experts whose opinion is most likely to disagree with that of the expert 
sought first.
Here, it is worth noting that several studies have also argued that decision support systems that identify disagreement between experts may help identify 
when a decision instance would benefit most from a second opinion~\citep{raghu2019direct, lim2021finding}. 
However, these studies do not focus on inferring other experts'{} opinions given an expert'{}s opinion on a decision instance and a set of features, 
as we do in our work.
%
%
%

\begin{figure*}[t]
        \centering
        \subfloat[Training dataset]{\includegraphics[width=0.4\textwidth]{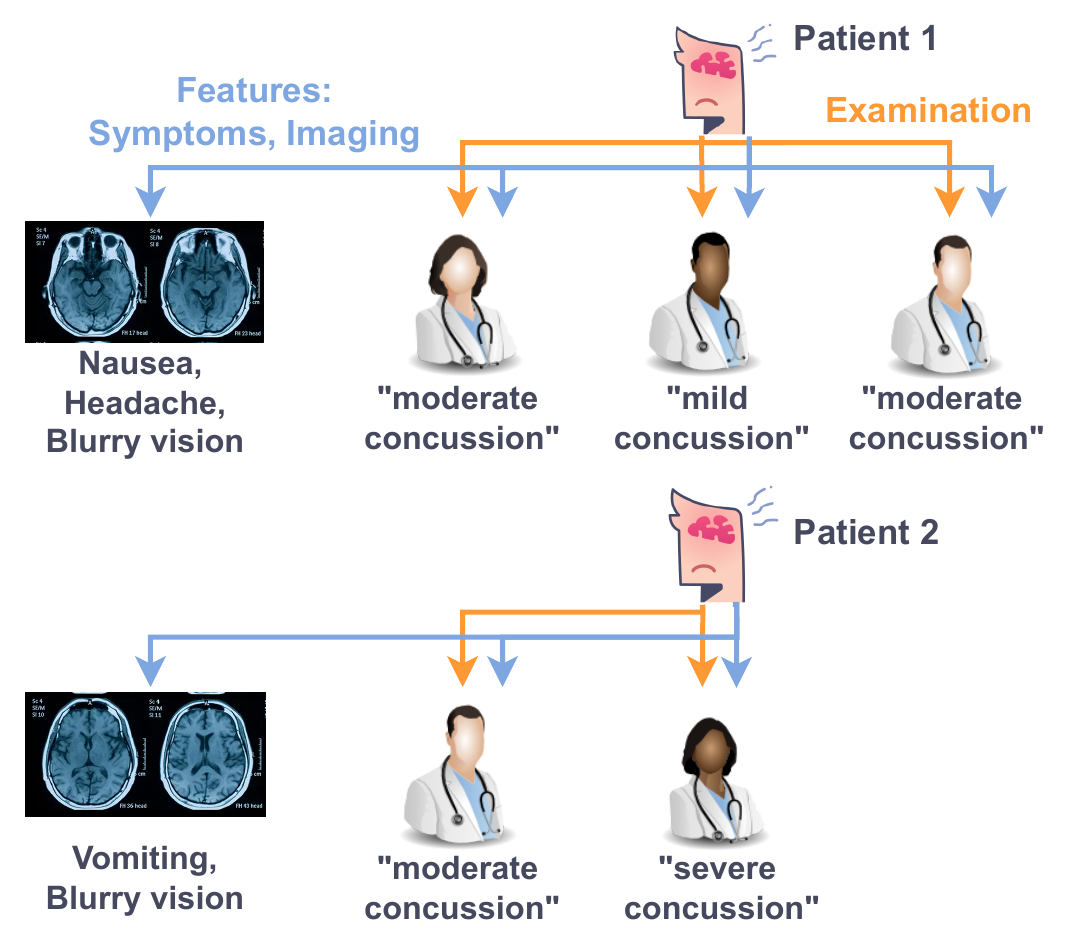}}
        \hspace{1mm}
        \subfloat[Use case]{\includegraphics[width=0.54\textwidth]{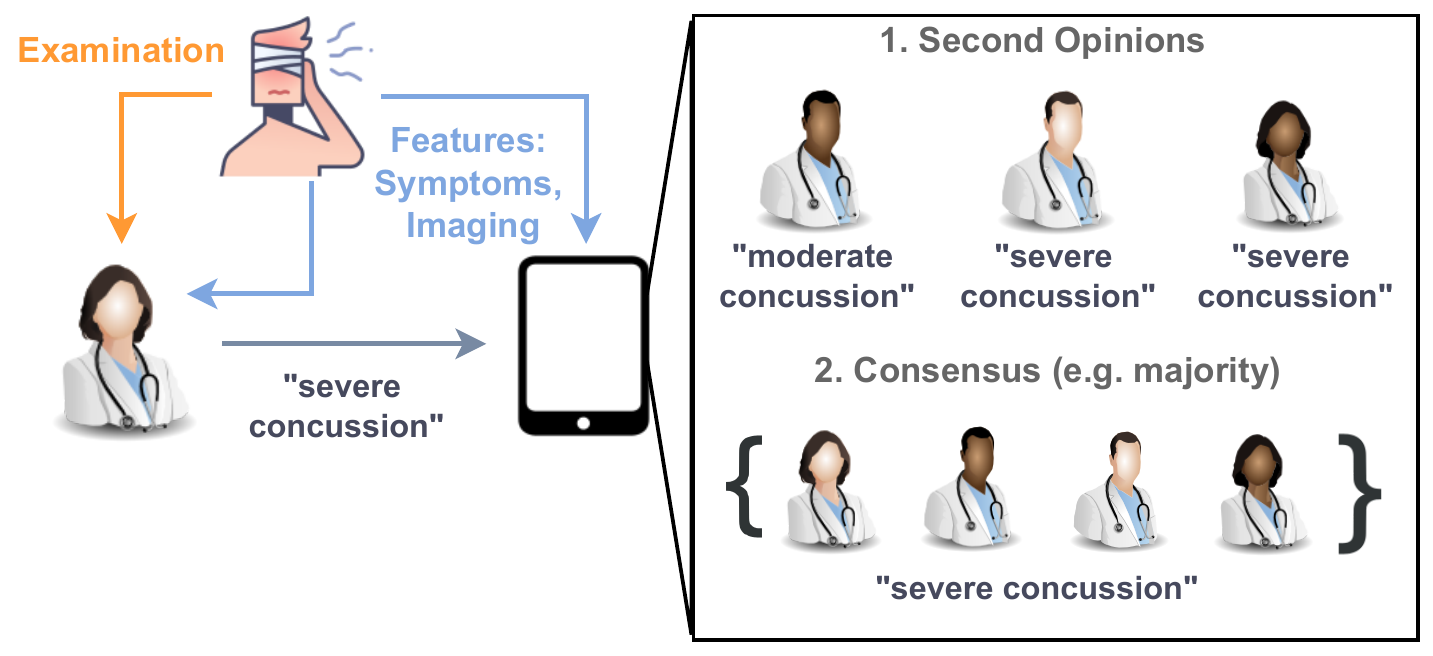}}
        \caption{An example of a training dataset and use case of our decision support system on a medical application.
        In panel (a), for each patient, multiple doctors assess the severity of a concussion on the basis of a set of features.
        In panel (b), given a doctor'{}s assessment of the severity of a concussion and a set of features, our decision support system infers other fellow doctors'{}
        assessment of the severity of the concussion.}
        %
\label{fig:intro_example}
\end{figure*}

%
%
More specifically, we consider a multiclass classification setting where, for each instance, experts form their opinions on their own (\ie, without 
communicating).\footnote{This setting fits a variety of real-world applications. For example, when a patient is diagnosed by multiple doctors, 
each doctor diagnoses the patient separately.} In this setting, each expert'{}s opinion reduces to a label prediction.
Then, our goal is to design decision support systems that, given an expert'{}s prediction on an instance with a set of features, are able to infer 
other experts'{} predictions about the same instance, as illustrated in Figure~\ref{fig:intro_example}.
%
%
%
%
%
To this end, one could resort to standard supervised learning. Under this perspective, for each instance, the given expert'{}s prediction would 
be just an additional feature about the instance.
Unfortunately, this would limit the applicability of the re\-sul\-ting supervised learning model to the unrealistic scenario where, for each 
possible pair of experts, we observe a sizeable number of instances where both experts made a prediction.
%
%
To circumvent this limitation, we look at the design of the above systems from the perspective of counterfactual inference.

\xhdr{Our contributions} We first show that, if experts form their opinions for each instance on their own, the underlying causal mechanism ge\-ne\-ra\-ting 
the experts'{} predictions needs to sa\-tis\-fy a certain set invariant property.
Moreover, we further show that any structural causal model satisfying the above set invariant property (in short, any SI-SCM) also satisfies two additional desirable 
properties: 
\begin{itemize}[noitemsep, topsep=0pt]
\item[(i)] there exists an equivalent SI-SCM where each expert'{}s predictions are generated by independent sub-mechanisms 
governed by a common (multidimensional) noise;
\item[(ii)] given an expert'{}s prediction on an instance with a set of features, the conditional interventional distribution and the counterfactual 
distribution of another expert'{}s predictions entailed by the SI-SCM are identical\footnote{Under the conditional interventional distribution, both experts have made a prediction but we only 
observe one of them. Under the counterfactual distribution, only one expert has made a prediction, which we observe.}. 
\end{itemize}
These properties suggest the following natural strategy to design and train SI-SCM based decision support systems.
In a first step, we can use interventional data about each expert---her predictions on a set of instances---to determine the structure of 
each sub-mechanism separately.
One can view this step as multiple independent supervised learning problems, one per expert.
In a second step, we can use a small amount of interventional data about multiple experts ma\-king predictions about a joint set of instances to 
characterize similarity across experts and factorize the noise governing the sub-mechanisms into a set of noise components. 
%
%
%
In a way, in this second step, we are adding a wrapper to the supervised learning models characterizing each expert'{}s sub-mechanism to be able to make 
counterfactual predictions about second opinions.

To implement the above strategy, we introduce a specific class of SI-SCMs based on the Gumbel-Max structural causal model~\citep{oberst2019} (in short, 
Gumbel-Max SI-SCM) and characterize similarity across pairs of experts using the concept of counterfactual stability\footnote{Counterfactual stability is, in general, 
an axiomatic requirement imposed to counterfactual distributions~\citep{oberst2019}. However, in SI-SCMs, it is verifiable from interventional data due to (ii), as 
shown in Theorem~\ref{thm:pcs_equivalence}.}.
%
In the Gumbel-Max SI-SCM, each expert'{}s submechanism is go\-verned by a Gumbel-Max noise variable
and submechanisms of similar experts may be governed by the same noise variable. 
%
Further, we show that the problem of uniquely asso\-cia\-ting each of these noise variables with disjoint sets of mutually similar experts given data can be formulated as a known
clique par\-tio\-ning problem, an NP-hard problem~\citep{grotschel1989cutting,grotschel1990facets},
and propose a simple randomized greedy algorithm with good performance.
Finally, we experiment with synthetic and real data comprising of $20{,}426$ expert predictions over $1{,}560$ natural images. 
The results on synthetic data show that our randomized greedy algorithm can successfully recover the disjoint sets of mutually similar experts underpinning a
specific Gumbel-Max SI-SCM from data.
The results on real data show that the (counterfactual) predictions provided by the Gumbel-Max SI-SCM are more accurate than those provided by its non-causal 
counterpart.
%
%
%

\xhdr{Further related work}
%
%
Predictions by different experts have been typically studied separately, \ie, without conditioning on an observed prediction by a given expert~\citep{dawid1979maximum, welinder2010online, 
guan2018said, kerrigan2021combining, straitouri2022provably}.
One could think of the observed prediction just as an additional feature when inferring other experts'{} predictions, however, this would limit the applicability of existing inference 
methods to scenarios where, for each pair of experts, we observe a sizeable number of instances where both experts made a prediction, as discussed previously.
More broadly, our work is not the first to use counterfactual reasoning in expert prediction~\citep{bica2020learning}. However, previous work has used counterfactual reasoning to 
quantify an expert'{}s preference over counterfactual outcomes rather than to infer other experts'{} predictions conditioning on a given expert'{}s prediction.

Counterfactual inference has a long and rich history~\citep{imbens2015causal}. However, it has mostly focused on estimating quantities related to the interventional 
distribution of interest such as, \eg, the conditional average treatment effect (CATE).
A few notable exceptions are by~\cite{oberst2019} and~\cite{tsirtsis2021counterfactual}, which use the Gumbel-Max SCM to reason about counterfactual distributions in Markov decision 
processes (MDPs), and by~\cite{lorberbom2021learning}, which introduces a parameterized family of causal mechanisms that generalize the Gumbel-Max SCM and are specifically-tuned to a distribution of observations and interventions of interest.
However, the Gumbel-Max structural causal model has not been used previously to reason about counterfactual expert predictions.

\vspace{-2mm}
\section{Preliminaries}
\label{sec:preliminaries}
\vspace{-2mm}
%


Given a set of random variables\footnote{We denote random variables with capital letters and realizations of random variables with lower case letters.} $\mathbf{X} = \{X_1, \dots X_n\}$, a structural causal 
model (SCM) $\M$ defines a complete data-generating process via a collection of assignments
\begin{equation*}
X_i = f_i(\mathbf{PA}_i, U_i),
\end{equation*}
where 
$\mathbf{PA}_i\subseteq \mathbf{X}\setminus X_i$ are the direct causes of $X_i$,
$\mathbf{F}=\{f_1, \dots, f_n\}$ are deterministic causal mechanisms, 
$\mathbf{U}=\{U_1, \dots, U_n\}$ are jointly independent noise random variables,
and $P(\mathbf{U})$ denotes the (prior) distribution of the noise variables.
%
%
Here, note that the noise variables $\mathbf{U}$ are the only source of stochasticity and, given an observational distribution $P(\mathbf{X})$, there always exists a distribution 
$P(\mathbf{U})$ and mechanisms $\mathbf{F}$ so that $P = P^{\M}$, where $P^{\M}$ is the distribution entailed by $\M$.

Two SCMs $\M$ and $\tilde{\M}$ over variables $\vecX$ and $\vecU$, with noise distribution $P(\mathbf{U})$ and mechanisms $\vecF$ and $\tilde{\vecF}$ respectively, are equivalent if, for all $i \in [n]$, it holds
that
\begin{equation*}
 x_i=f_i(\mathbf{pa}_i,u_i) \Longleftrightarrow x_i=\tilde{f}_i(\mathbf{pa}_i,u_i).
\end{equation*}
for any realization $\mathbf{PA}_i=\mathbf{pa}_i$ and $P(\mathbf{U})$-almost every $u_i$.
\footnote{ 
$P(\mathbf{U})$-almost everywhere means that the set of noise realizations $\mathcal{U’}$ for which the property does not hold has probability zero under the distribution $P(\mathbf{U})$, i.e., $P(\mathbf{U} \in \mathcal{U’})=0$.
}
%
%

%
Given a SCM $\M$, an atomic intervention $\Ical$ corresponds to assigning a fixed value to a variable. 
%
For example, let $\Ical = \text{do}[X_i=x]$ be the intervention that assigns value $x$ to variable $X_i$, then the intervened SCM $\M^{\Ical}$ does not assign the value of $X_i$ according to $f_i(\mathbf{PA}_i,U_i)$ 
but assign it to a fixed value $x$. The interventional distribution entailed by the intervened SCM is denoted $P^{\M\,;\,\Ical}$.
Furthermore, given the (possibly partial) observation $\mathbf{X}=\mathbf{x}$,  we can also define a modified SCM $\M_{\Xb=\xb}$ where the
noise variables $\Ub$ are distributed according to the posterior distribution $P(\Ub \given \Xb=\xb)$.
Then, we can view a counterfactual statement as an intervention $\Ical$ in the SCM $\M_{\Xb=\xb}$ and denote the counterfactual distribution 
entailed by the counterfactual SCM $\M^{\Ical}_{\Xb=\xb}$ as $P^{\M \given \mathbf{X}=\mathbf{x}\,;\,\Ical}$. 
The Gumbel-Max SCM is a specific class of SCM in which the causal mechanism for a random categorical variable $V$ is defined as
%
%
\begin{equation}
	f_v(\mathbf{PA}, \mathbf{U}) := \argmax_j\{\log P(V=j \given \mathbf{PA}) + U_j\}
	\label{eq:gumbelscm}
\end{equation}
and each noise variable $U_j \sim \text{Gumbel}(0,1)$.
%
%
Here, note that the interventional distribution $P^{\Mcal\,;\,do[\mathbf{PA} = \mathbf{pa}]}(V)$ entailed by a Gumbel-Max SCM $\Mcal$ is exactly $P(V \given \mathbf{PA} = \mathbf{pa})$.


\vspace{-2mm}
\section{Counterfactual Inference of Second Opinions}
\label{sec:problem}
\vspace{-2mm}
%
%

We consider a multi-class classification task where, for each instance, a human expert $h \subseteq \Hcal$ makes a label prediction 
$y_h \in \Ycal = \{1, \ldots, k\}$ based on multiple sources of information, which are (imperfectly) summarized by a feature vector $x \in \Xcal$. 
Here, we assume that experts make predictions on their own (\ie, without communicating with each other) and the assignment of experts 
to instances is independent of the identity of the instances and their feature vectors.
%
%
Then, our goal is to design an automated decision support system that, 
given a prediction $y_h$ from an expert $h$ about an instance summarized by a feature vector $x$, 
is able to infer what prediction $y_{h'}$ another expert $h' \neq h$ would have made about the \emph{same} instance if she had 
been asked.
Here, note that two different instances may be (imperfectly) summarized by the same feature vector $x$, however, we are interested
in a counterfactual prediction about the \emph{same} instance.

Our starting point is to view the above counterfactual statement as an intervention in a particular counterfactual SCM. More 
specifically, let $\Mcal$ be a SCM defined by the assignments
\begin{equation} \label{eq:scm}
\vecY = f_\vecY(X, Z, U), \quad Z = f_{Z}(V), \quad \text{and}  \quad X = f_{X}(W)
\end{equation}
where $U$, $V$ and $W$ are (multidimensional) independent noise variables, $f_{\vecY}$, $f_{Z}$ and $f_{X}$ are given deterministic causal mechanisms (or functions),
and $\Yb = (Y_h)_{h \in Z}$ are the predictions by a set of human experts $Z \subseteq \Hcal$.
Then, we can express the above counterfactual statement as an intervention $\Ical = do[Z = \{h'\}]$ in the counterfactual SCM 
$\Mcal_{X = x, Z = \{h\}, \Yb = y_h}$ and, to infer the label prediction $y_{h'}$, we just need to resort to the counterfactual 
distribution $P^{\Mcal \given X = x, Z = \{h\}, \Yb = y_{h} \,;\, do[Z = \{h'\}]}(\Yb)$.
%
%
%

At this point, one may argue that, even if we find a noise distribution $P(U)$ and a function $f_{\Yb}$ under which the conditional distribution 
$P^{\Mcal}(\Yb \given X)$ is a good fit for observed historical predictions by experts, we would be unable to validate how accurate our 
counterfactual label predictions are using data. 
In general, this is true since counterfactual reasoning lies within level three in the ``ladder of causation''~\citep{pearl2009causality}. In this context, 
previous work resorts instead to axiomatic assumptions about the causal mechanism of the world~\citep{oberst2019, tsirtsis2021counterfactual, noorbakhsh2021counterfactual}. 
In our setting, this would reduce to specifying how differences across experts may have lead to a different prediction while holding ``everything else'' fixed.
However, in what follows, we will show that, if experts do not communicate with each other, the above SCM satisfies a set invariance property that sur\-pri\-singly
implies that the above counterfactual distribution coin\-cides with an interventional conditional distribution. This enables a data-driven design and validation of our SCM based decision support system.

\vspace{-2mm}
\section{Relating the Counterfactual and Interventional Worlds}
\label{sec:cscm}
\vspace{-2mm}

%
%
To build some intuition on the reasons why, if experts do not commu\-ni\-cate, certain type of counterfactual and interventional distributions are identical, we start 
with a simple example. 
Let $h, h' \in \Hcal$ be two different experts and consider the following two questions:
\begin{enumerate}
	\item Both experts have made a label prediction about an instance (\ie, $Z=\{h, h'\}$) but we only observe the prediction $Y_{h} = c$ made by $h$, what is the
	prediction made by $h'$? 
		\label{it:q1}
	\item One of the experts has made a label prediction $Y_{h} = c$ about an instance (\ie, $Z=\{h\}$) and we observe it, what would the prediction made by $h'$ be 
	if she had made a prediction?
		\label{it:q2}
\end{enumerate} 
The first question is of conditional nature while the second is a counterfactual one.
%
In general, the answer to both questions may differ, for example, if experts 
influence each other'{}s predictions by sharing and discussing their opinions 
in the first case. 
However, if experts do not commu\-ni\-cate, the answer to both questions should be identical.
More formally, the following conditional interventional distribution and counterfactual distribution of the expert should be equal:
%
\begin{equation}
	P^{\M \,;\, \text{do}[Z=\{h,h'\}]}(Y_{h'} \given X=x, Y_{h}=c)
	= P^{\M \given X=x, Z=\{h\},\Yb=c \,;\, \text{do}[Z=\{h'\}]}(\Yb).
	\label{eq:samedistr_example}
\end{equation}
%
More generally, we will now show that, if each expert forms their opinion on their own, the above equality is a direct consequence of a set invariance property satisfied 
by the SCM defined in Eq.~\eqref{eq:scm}.

\xhdr{Set Invariant SCMs (SI-SCMs)}
%
If experts do not communicate before making a prediction and hence are unaware and unaffected of other experts' opinions,
%
%
the me\-cha\-nism $f_{\vecY}$ has a set in\-va\-riant value over expansions (supersets) of $Z$. 
For example, consider one single expert $h$ has made a prediction $f_{\vecY}(x,\{h\},u)=c$ about a specific instance. 
%
%
Then, one can conclude that, if instead of a single expert, a set of experts $\zeta \subseteq \Hcal$ such that $h \in \zeta$ would
have made predictions about the same instance (\ie, $x$ and $u$ does not change), 
expert $h$ would have made the same prediction, \ie, $( f_{\vecY}(x,\zeta,u) )_h = c$. 
%
More formally, we define the set invariance property as follows:
\begin{definition}[Set Invariance]\label{prop:set invariant}
	A mechanism $f_{\vecY}$ for variable $\vecY$ is set invariant with respect to $Z$ if, for any two realizations $Z=\zeta$ and $Z=\zeta'$ such that 
	$\zeta\subseteq \zeta'$, it holds~that
	\begin{equation*}
	f_{\vecY}(x, \zeta, u) = (f_{\vecY}(x,\zeta',u))_{\zeta}  \quad \text{ for all } x \in \X, u \in \U\ .
	\end{equation*}
	A SCM $\M$ with such a mechanism is set invariant for $\vecY$.
\end{definition}

%
%
	A set-invariant SCM (SI-SCM) for $Y$ can be constructed by expressing the causal mechanism $f_{\vecY}$ with submechanisms $f_{Y_h}$ governed by a common noise variable:\footnote{All proofs can be found in Appendix~\ref{app:awesomeproofs}}
\begin{theorem}\label{thm:SCMsetinvariant}
	Any SCM $\M$ with mechanism $f_{\vecY}$ of the form $f_{\vecY}(X,Z,U)= (f_{Y_h}(X,U))_{h \in Z}$, where $f_{Y_h} \colon \X \times \U \to \Y$ are arbitrary functions, 
	is set invariant for $\vecY$.
\end{theorem}
%
In fact, the following theorem shows that the class of SCMs with separate submechanisms for $Y_h$ and a shared noise variable $U$ is not only a subclass but completely defines the class of SI-SCMs for $\vecY$. 
Thus, any correlation between experts' predictions is caused by the common noise and features but not the causal mechanism.
\begin{theorem}\label{thm:SCMequivalence}
 For any SI-SCM $\M$, there exists an equivalent SI-SCM $\M'$ with causal mechanism 
$f'_{\vecY}(X,Z,U) = (f'_{Y_h}(X,U))_{h \in Z}$ where for $h \in Z$
 %
\begin{equation*}
f'_{Y_h}(X,U) := (f_{\vecY}(X,\{h\},U))_{h\in Z}.
\end{equation*}
\end{theorem}
%
Here, we would like to emphasize that, if the mechanism $f_{\vecY}$ of an SCM is not explicitly decoupled into submechanisms governed by the same noise, it may be challenging to check whether an arbitrary SCM is set invariant.
For arbitrary SCMs, Theorem~{\ref{thm:SCMsetinvariant}} can not be applied directly and Theorem~{\ref{thm:SCMequivalence}} does not tell us how to verify that an equivalent SCM exists.
However, it tells us that the mechanism of a set invariant SCM can be decoupled and simplified. It would be interesting to develop methods to check for set invariance for arbitrary SCMs in future
work.
%

\xhdr{Equality between the counterfactual distribution and the conditional interventional distribution}
%
%
Returning to our simple motivational example, note that, if a SCM is set invariant, the answers to the counterfactual and the conditional questions~\ref{it:q1} and~\ref{it:q2} are 
the same as long as the noise $u \sim P(U \given X=x, Y_{h}=c)$ is the same.
In particular, for question~\ref{it:q1}, the answer is $Y_{h'}=(f_{\vecY}(x,\{h,h'\},u))_{h'}$, for question~\ref{it:q2}, the answer is $Y_{h'}=f_\vecY(x,\{h'\},u)$, and since $f_{\vecY}$ is set invariant,
both answers are equal. 
More generally, for arbitrary sets of experts, we can easily conclude that equality holds if and only if $f_{\vecY}$ is set invariant. 

Next, to show that, if a SCM is set invariant, then the equa\-li\-ty of distributions in Eq.~\eqref{eq:samedistr_example} holds, we first present a more 
general theorem that states that, if we expand the set of experts who make predictions, the corresponding interventional distribution of $\vecY$
does not change:
%
\begin{theorem}\label{thm:equalprop}
	%
	Let SCM $\M$ be set invariant for $\vecY$. Then, for any $\zeta, \zeta' \in \Hcal$ such that $\zeta \subseteq \zeta'$,
	it holds that
	\begin{equation*}
		P^{\M \,;\, \text{do}[Z=\zeta]}(\vecY=\vecy \given X)
		= P^{\M \,;\, \text{do}[Z=\zeta']}((\vecY)_{\zeta}=\vecy\given X)
	\end{equation*}
	for any $\vecy \in \Y^{|\zeta|}$ where $(\vecY)_{\zeta}$ denotes the predictions by the experts in the 
	subset $\zeta \subseteq \zeta'$.
\end{theorem}
The above theorem is straight forward to show using that, due to the set invariance property, the prediction values of mechanism $f_{\vecY}$ for $(x, \zeta, u)$ are equal to the values for $(x,\zeta',u)$ for experts 
in $\zeta$ and, due to the independence between the noise and the intervention, the noise distribution does not change. 
A direct conclusion 
is that, no matter how many experts make predictions, the conditional interventional distribution of a single expert's prediction does not change, 
as formalized by the following corollary:
%
\begin{corollary}\label{cor:equalprop_h}
Let SCM $\M$ be set invariant for $\vecY$. Then, for any $h \in \Hcal$ and $\zeta \subseteq \Hcal$ such that $h \in \zeta$, it 
holds that
        \begin{equation*}
        P^{M \,;\, do[Z=\{h\}]}(Y_h \given X) = P^{M \,;\, do[Z=\zeta]}(Y_h\given X).
        \end{equation*}
\end{corollary}

Similarly, we can derive the desired equality between the counterfactual distribution and the conditional interventional distribution by using the set invariance of mechanism $f_{\vecY}$ and the fact that the noise 
distribution changes equally in both scenarios. More formally, we have the following corollary:
\begin{corollary}\label{cor:equalcounterfactuals}
Let SCM $\M$ be set invariant for $\vecY$. Then, for any $h,h' \in \Hcal$ and $\zeta \subseteq \Hcal$ such that $h,h' \in \zeta$, it holds that
\begin{equation*}
	P^{\M\given X=x, Z=\{h\},\vecY=c \,;\, \text{do}[Z=\{h'\}]}(\vecY)
	= P^{\M \,;\, \text{do}[Z=\zeta]}(Y_{h'}\given X=x, Y_{h}=c)
\end{equation*}
for any $x \in \X$ and $c \in \Y$.
\end{corollary}
%
\xhdr{Remark} While we have introduced the notion of set invariance for SCMs in the context of inferring second opinions, we believe it may be of independent interest since, generally speaking, it allows 
us to identify counterfactual distributions from interventional data.


\vspace{-2mm}
\section{Characterizing Mutually Similar Experts} 
\label{sec:pcs}
\vspace{-2mm}
%
%
	Given a SI-SCM model $\Mcal$ where each expert'{}s predictions $Y_h$ are generated by a sub-mechanism $f_{Y_h}$, our goal in 
this section is to characterize mutually similar experts. 
%
Later on, this will help us factorize the noise $U$ gover\-ning the sub-mechanisms $f_{Y_h}$ underpinning the model into a set of independent 
noise components and uniquely associate each of these noise components with disjoint sets of mutually similar experts given data.

To this end, we first start by characterizing similarity between a pair of experts $h, h' \in \Hcal$. 
To this end, we resort to the recently introduced notion of counterfactual sta\-bi\-li\-ty~\citep{oberst2019}. More specifically, we argue that two experts 
$h$ and $h'$ are \emph{similar} if $\M$ satisfies counterfactual stability for $h, h'$ with respect to $\vecY$.
\begin{definition}[Counterfactual stability]
\label{def:counterfactual-stability}
A SCM $\M$ satisfies counterfactual stability for $h, h'$ with respect to $\vecY$ if, for all $\zeta, \zeta' \subseteq \Hcal$ such that $h \in \zeta$ and 
$h' \in \zeta'$ and for all $c' \neq c$, the condition
\begin{equation*}
\frac{P^{\M \,;\, \text{do}[Z = \zeta']}(Y_{h'} = c \given X)}{P^{\M \,;\, \text{do}[Z = \zeta]}(Y_h = c \given X)} \geq
\frac{P^{\M \,;\, \text{do}[Z = \zeta']}(Y_{h'} = c' \given X)}{P^{\M \,;\, \text{do}[Z = \zeta]}(Y_h = c' \given X)}
\end{equation*}
	implies that $P^{\M \given X, Z=\zeta, Y_{h} = c \,;\, \text{do}[Z = \zeta']}(Y_{h'} = c') = 0$, where $Y_{h} = c$ is the observed outcome
under $\text{do}[Z = \zeta]$.
\end{definition}
For example, consider a scenario where a doctor needs to decide what treatment option---surgery ($Y=0)$, radiation ($Y = 1$) or chemotherapy ($Y = 2$)---will be 
more beneficial for a patient with a tumor, imperfectly summarized by a feature vector $x$.
%
%
Assume doctor $h$ decides the most beneficial option is surgery, \ie, $Y_h = 0$, and we know that, for patients with similar $x$, doctor $h'$ is generally more likely 
to operate and less likely to resort to therapy than doctor $h$.
Then, if doctors $h$ and $h'$ are similar, as defined in Definition~\ref{def:counterfactual-stability}, we expect doctor $h'$ would have also decided the most be\-ne\-fi\-cial option
is surgery for the given patient, if consulted, \ie, $Y_{h'} = 0$. 
Here, whenever two doctors $h$ and $h'$ are \emph{not} similar, one could argue that it is because they weigh any (hidden) factor specific to the patient at hand 
differently\footnote{In general, note that similarity between experts does not always deterministically enforce the observed expert'{}s prediction on the counterfactual prediction. In the
example above, this happens because the inequality in Def.~\ref{def:counterfactual-stability} holds for the two remaining label values.
Rather, it allows us to identify experts with different decision making criteria.}.

%
%
%
Unfortunately, in general, we cannot use data to verify if two experts $h$ and $h'$ are similar. This is because our no\-tion of similarity relies on a counterfactual distribution, 
$P^{\M \given Y_{h} = c \,;\, \text{do}[Z = \zeta']}$, and counterfactual reasoning lies within level three in the ``ladder of causation''~\citep{pearl2009causality}. 
However, we will now define a notion of conditional sta\-bi\-li\-ty that is verifiable using interventional data and, in the case of SI-SCMs, is both a sufficient and necessary 
condition for counterfactual stability---if conditional sta\-bi\-li\-ty holds, we can conclude that two experts are similar.
\begin{definition}[Conditional stability]
	A SCM $\M$ satisfies conditional stability for two experts $h, h' \in \Hcal$ with respect to $\vecY$ if, for all $\zeta \subseteq \Hcal$ such 
	that $h, h' \in \zeta$ and for all $c' \neq c$, the condition
	\begin{equation}
		\frac{P^{\M \,;\, \text{do}[Z = \zeta]}(Y_{h'} = c \given X)}{P^{\M \,;\, \text{do}[Z = \zeta]}(Y_h = c \given X)} \geq
		\frac{P^{\M \,;\, \text{do}[Z = \zeta]}(Y_{h'} = c' \given X)}{P^{\M \,;\, \text{do}[Z = \zeta]}(Y_h = c' \given X)}
		\label{eq:conditional-stability}
	\end{equation}
	 implies that 
	$P^{\M;do[Z=\zeta]}(Y_{h'}=c' \given X, Y_{h} = c)=0$.
\end{definition}
Here, note that, for SI-SCMs, we only need to verify the condition in Eq.~\eqref{eq:conditional-stability} for the sets $\zeta = \{h\}$ and $\zeta = \{h'\}$ because 
no matter how many experts make predictions, the conditional interventional distributions in Eq.~\eqref{eq:conditional-stability} do not change, as shown in 
Corollary~\ref{cor:equalprop_h}.
Then, the following Theo\-rem formalizes the equivalence between conditional and counterfactual stability:
%
\begin{theorem}\label{thm:pcs_equivalence}
	Let SCM $\M$ be set invariant for $\vecY$. Then, $\M$ satisfies counterfactual stability for $h, h' \in \Hcal$ with 
	respect to $\vecY$ iff it satisfies conditional stability.
\end{theorem}
Once we have a notion of similarity between pairs of experts that we can verify from data, we can characterize groups of mutually similar
experts. In this context, it will be useful to introduce the following notion of pairwise counterfactual stability (in short, PCS), which extends 
counterfactual stability to groups of experts $\zeta \subseteq \Hcal$ of arbitrary size.
\begin{definition}[Pairwise Counterfactual Stability]\label{def:pcs_def}
	A SCM $\M$ satisfies pairwise counterfactual stability for a group of experts $\zeta \subseteq \Hcal$ with respect to $\vecY$ if
	it satisfies counterfactual stability for any $h, h' \in \zeta$.
\end{definition}
Similarly as in the case with a pair of experts, one can also define pairwise conditional stability and it imme\-dia\-te\-ly follows\-
from Theorem~\ref{thm:pcs_equivalence} that, for SI-SCM, pairwise conditional and counterfactual stability are equivalent, 
as formalized by the following Corollary.
\begin{corollary}\label{cor:pcs_equivalence}
	Let SCM $\M$ be set invariant for $\vecY$. Then, $\M$ satisfies pairwise counterfactual stability for $\zeta \in \Hcal$ with 
	respect to $\vecY$ iff it satisfies pairwise conditional stability.
\end{corollary}

\vspace{-2mm}
\section{Gumbel-Max SI-SCM}
\label{sec:gumbel}
\vspace{-2mm}


In this section, we build upon our theoretical results to develop the Gumbel-Max SI-SCM, a new class of SI-SCM based on the 
Gumbel-Max SCM. 


%

Given a set of experts $\Hcal$, the Gumbel-Max SI-SCM partitions $\Hcal$ into disjoint sets of experts $\Psi=\{\psi\}_{\psi \in \Psi}$, 
as defined in Section~\ref{sec:pcs}, and associate all experts within each set to the same multidimensional noise variable.
%
%
More formally, the Gumbel-Max SI-SCM is defined as follows:
%
%
%
\begin{definition}[Gumbel-Max SI-SCM] 
%
%
The Gumbel-Max SI-SCM $\M(\Psi)$ is a specific class of SCM in which the causal mechanism for $\vecY$ is defined as
%
%
\begin{equation*}
	f_{\vecY}(X, Z, U) = (f_{Y_h}(X, U))_{h \in Z},
\end{equation*}
with
%
%
\begin{equation*}
	f_{Y_h}(X, U_{\psi(h)}) = \argmax_{c \in \Y}\{ \log P(Y_h=c \given X) + U_{\psi(h), c}\},
\end{equation*}
where $\psi(h) \in \Psi$ denotes the subgroup expert $h$ belongs to and each noise variable $U_{\psi(h), c} \sim \text{Gumbel}(0,1)$.
%
\end{definition}
\vspace{-2mm}
By definition, the Gumbel-Max SI-SCM $\M(\Psi)$ is set invariant for $\Y$ and,
for any $\zeta \subseteq \Hcal$ and $h \in \zeta$, it holds that
$P^{\M(\Psi);\text{do}[Z=\zeta]}(Y_h \mid X) = P(Y_h \mid X)$.
Moreover, all experts within each group $\psi \in \Psi$ are mutually similar, as formalized by the following Theorem:
\begin{theorem} \label{thm:gumbel}
	The Gumbel-Max SI-SCM $\M(\Psi)$ satisfies pairwise counterfactual stability (PCS) for each group $\psi \in \Psi$ with respect to $\vecY$. 
\end{theorem}
\vspace{-2mm}
Finally, note that, for $\Psi = \Hcal$, the Gumbel-Max SI-SCM reduces to the original Gumbel-Max SCM defined 
in Eq.~\ref{eq:gumbelscm}.
Therefore, one can view the Gumbel-Max SI-SCM as a generalization of the original Gumbel-Max SCM where,
instead of a single multidimensional noise variable $U$ for all $h \in \Hcal$, one has several noise variables $U_{\psi}$,
one per group.

%
%

\begin{figure}[t]
	\centering
\begin{minipage}{.5\textwidth}
        \centering
                \includegraphics[width=1\textwidth]{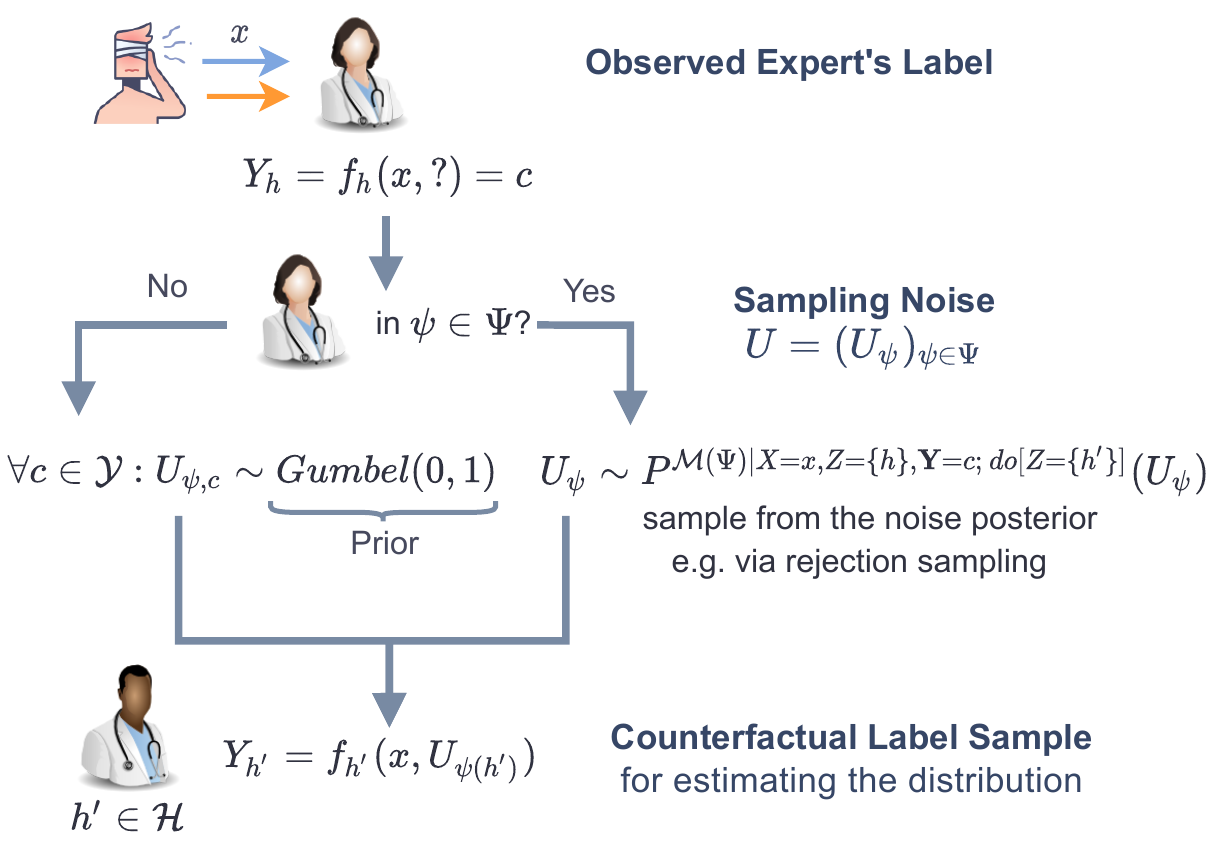}
        \vspace{-2mm}
        \caption{ Illustration of the counterfactual sampling of experts' predictions with the Gumbel-Max SI-SCM $\M(\Psi)$.
        }
        \vspace{-2mm}
\label{fig:flowdiagram_inference}
\end{minipage}
	\hspace{1em}
\begin{minipage}{.46\textwidth}
        \centering
        \includegraphics[width=0.7\textwidth]{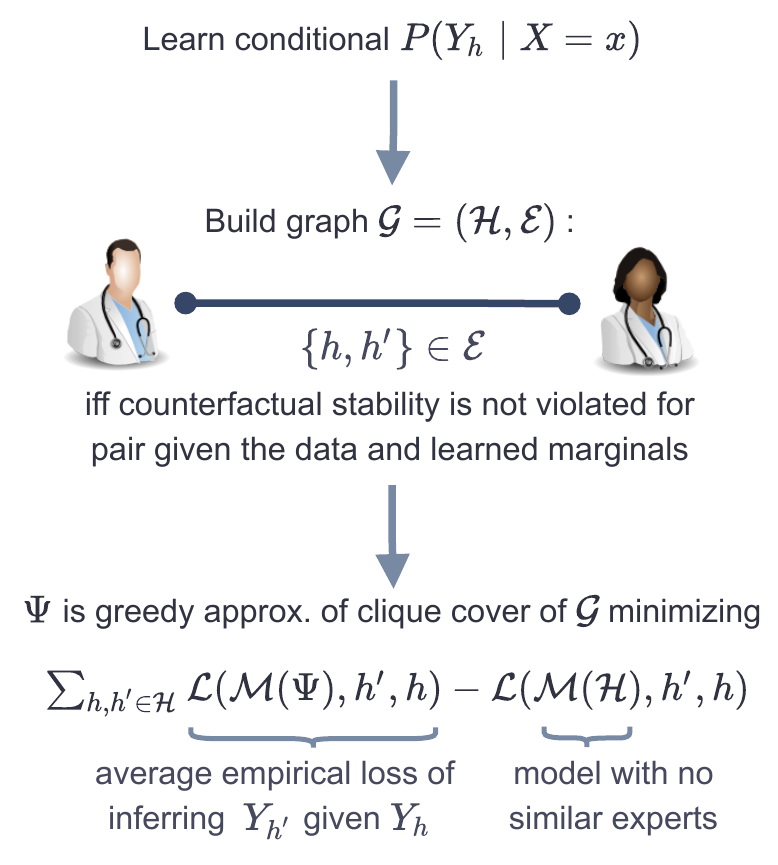}
        \vspace{-2mm} 
        \caption{Partitioning experts into mutually similar groups $\Psi$.
        }
        \vspace{-2mm}
\label{fig:flowdiagram_training}
\end{minipage}
\end{figure}

%
\xhdr{Estimating counterfactual distributions} 
Given a prediction $Y_h=c$ by an expert $h$, we can compute an unbiased finite sample Monte-Carlo estimator of the counterfactual
distribution for the prediction $Y_{h'}$ of another expert $h' \neq h$, \ie, $P^{\Mcal(\Psi) \given X = x, Z = \{h\}, \Yb = y_{h} \,;\, \text{do}[Z = \{h'\}]}(\Yb)$,
as follows:
\vspace{-2mm}
\begin{equation} \label{eq:cfc_distr_estimator}
P^{\Mcal(\Psi) \given X = x, Z = \{h\}, \Yb = y_{h} \,;\, do[Z = \{h'\}]}(\Yb) 
	\approx \frac{1}{T} \sum_{t \in T} \mathbbm{1}{[c=f_{Y_{h'}}(x,\mathbf{u}_t)]} 
\end{equation}
%
where $\mathbf{u}_1, \ldots, \mathbf{u}_T$ are samples from the posterior distribution $P^{\Mcal(\Psi) \given X = x, Z = \{h\}, \Yb = y_{h} \,;\, do[Z = \{h'\}]}(U_{\psi(h')})$
of the noise variable $U_{\psi(h')}$. 
Here, we can use an efficient procedure to sample from the above noise posterior distribution, described elsewhere~\citep{oberst2019,maddison2015a}.
Moreover, note that, if $h \notin \psi(h')$, the posterior distribution coincides with the prior $P^{\Mcal(\Psi)}(U_{\psi(h')})$.
%
	We summarized the sampling procedure is depicted in 
Figure~\ref{fig:flowdiagram_inference}.
%

%
%
%


\xhdr{Partitioning experts into mutually similar groups}
In the Gumbel-Max SI-SCM $\Mcal(\Psi)$, for each expert $h \in \Hcal$, we can estimate the conditional distribution $P(Y_h \given X)$ 
using any machine learning model trained using historical predictions made by the expert $h$. 
However, to fully define $\Mcal(\Psi)$, we need to partition the set of experts $\Hcal$ into disjoint sets of experts $\Psi$ given a small
amount of historical data about multiple experts making predictions about a joint set of instances. To this end, we proceed as follows.

First, we look for violations of the conditional stability condition throughout the historical data. 
Whenever there exists a sample for which the predictions by two different experts $h$ and $h'$ violate conditional 
stability\footnote{A violation occurs whenever Eq.~\eqref{eq:conditional-stability} holds but we observe $Y_h=c$ and $Y_{h'}=c'$.}, 
we conclude that $h$ and $h'$ cannot belong to the same group $\psi$.
Further, we also conclude that any pair of experts whose predictions did not violate conditional stability and were at least once 
observed for the same sample \emph{can} be similar.
However, since conditional stability is not a transitive property, there may be multiple valid partitions $\Pcal = \{\Psi\}$ of the experts into 
disjoint sets that are 
consistent with the above conclusions.
To decide among them, we would like to pick the partition $\Psi \in \Pcal$ under which the counterfactual distributions $P^{\Mcal(\Psi) \given X = x, Z = \{h\}, \Yb = y_{h} \,;\, do[Z = \{h'\}]}(\Yb)$ 
provide the best goodness of fit.  
More formally, we would like to solve the following mini\-mi\-za\-tion problem:
%
%
%
\vspace{-2mm}
\begin{equation} \label{eq:optproblem}
	\underset{\Psi \in \Pcal}{\text{minimize}} \quad \sum_{h,h' \in \Hu} \Loss(\M(\Psi),h',h) - \Loss(\M(\Hu),h',h), 
\end{equation}
%
where $\Loss(\M(\cdot),h',h)$ denotes an average (empirical) loss whenever we observe $Y_{h}$ and infer the label prediction $Y_{h'}$ using the 
counterfactual distribution $P^{\Mcal(\cdot) \given X = x, Z = \{h\}, \Yb = y_{h} \,;\, do[Z = \{h'\}]}(\Yb)$. 
Here, we measure goodness of fit in terms of average loss reduction with respect to the counterfactual distributions entailed by the causal model 
$\Mcal(\Hcal)$ because this will allow us to reduce the number of pairs $(h, h')$ we need to consider.
	The step-wise approach for obtaining $\Psi$ is summarized in
Figure~\ref{fig:flowdiagram_training}.

Next, we formulate the above problem as a known clique partitioning problem~\citep{grotschel1989cutting,grotschel1990facets}.
%
%
%
%
%
%
More specifically, let $\Gcal = (\Hu,\mathcal{E})$ be an undirected graph where, if $\{h, h'\} \in \Ecal$, then $h$ and $h'$ \emph{can} be similar, 
as concluded from the data. 
%
%
%
Then, it readily follows that finding a partition $\Psi$ of $\Hu$ is equivalent to finding a clique cover for $\Gcal$\footnote{$\Psi$ is a clique cover
for $\Gcal$ iff $\Psi$ is a partition of $\Hcal$, \ie, $\bigcup_{\psi \in \Psi} \psi = \Hcal$ and $\psi \cap \psi' = \emptyset$ for all $\psi,\psi' \in \Psi$, and vertices 
in $\psi \in \Psi$ form a clique in $\Gcal$.}.
Now, 
let the weight $w(h, h')$ of each edge $\{h, h'\} \in \Ecal$ be given by: 
\begin{equation*}
	w(h,h') = \Loss(\M(\Psi),h,h') - \Loss(\M(\Hu),h,h') 
	+ \Loss(\M(\Psi),h',h) - \Loss(\M(\Hu),h',h)
\end{equation*}
%
%
%
Then, we can rewrite the minimization problem defined in Eq.~\ref{eq:optproblem} as:
%
%
%
%
%
\begin{equation}
	\begin{split} 
		\underset{\Psi}{\text{minimize}} & \quad \sum_{\psi \in \Psi} \quad \sum_{h, h' \in \psi} w(h,h')
		\\
		\text{subject to} &\quad \Psi \text{ is a clique cover for } \Gcal, 
		\label{eq:optproblem_graph}
	\end{split}
\end{equation}
where note that we only need to consider pairs of experts $h, h' \in \psi$ because, otherwise, $w(h, h') = 0$ since the corresponding 
counterfactual distributions entailed by $\Mcal(\Psi)$ and $\Mcal(\Hcal)$ coincide.
The minimization problem given by Eq.~\eqref{eq:optproblem_graph} is a known clique partitioning problem (CPP)
\footnote{In most of the literature, 
the problem is defined for complete graphs. However, for arbitrary graphs, one can simply include the missing edges and assign positive infinite weights so that 
they are not included in a solution~\citep{brimberg2017solving}},
for which the decision problem of CPP for arbitrary weights is NP-Hard~\citep{grotschel1989cutting,grotschel1990facets}.
However, we found that a simple
randomized greedy algorithm works well in our setting, as shown in Figure~\ref{fig:results_synthetic} in Appendix~\ref{sec:synthetic}. Refer 
to Appendix~\ref{sec:algorithm} for more details about the algorithm.

%

%
%
%



\vspace{-2mm}
\section{Experiments on Real Data}
\label{sec:real}
\vspace{-2mm}
In this section, we compare the performance of the proposed Gumbel-Max SI-SCM at inferring second opinions against several competitive baselines using a 
dataset with real expert predictions over natural images. Appendix~\ref{sec:synthetic} contains addi\-tio\-nal experiments on synthetic data where we assess the 
performance of Algorithm~\ref{alg:greedyalg} at recovering groups of mutually similar experts on synthetic data.\footnote{To facilitate research in this area, we release an open-source implementation of our code at \href{https://github.com/Networks-Learning/cfact-inference-second-opinions}{https://github.com/Networks-Learning/cfact-inference-second-opinions}.}
%

\xhdr{Data description and experimental setup}
We experiment with the dataset CIFAR-10H~\citep{peterson2019human}, which contains $10{,}000$ images taken from the test set of the standard dataset CIFAR-10~\citep{krizhevsky2009learning}. 
Each of these images belongs to $n = 10$ classes and contains label predictions from approxi\-ma\-te\-ly $50$ human annotators. In total, the images are annotated 
by $2{,}571$ different human annotators (from now on, experts).\footnote{The dataset CIFAR-10H 
is one of the only publicly available datasets containing a relatively large number of samples with multiple label predictions by different experts per sample, necessary 
to train the proposed Gumbel-Max SI-SCM.
However, since our methodology and theoretical results are rather general, our model may also be useful in other applications.}
%
%
%
Since the classification task is relatively easy for humans, there are many images ($\sim$$35$\%) in which there is full agreement between experts---all experts make 
the same label prediction.
Here, motivated by the empirical observation that, in medical diagnosis, there is typically a $20\%$ per-instance disagreement among experts~\citep{van2017extent, elmore2015diagnostic}, we filter out the
above mentioned images in which there is full agreement.
%
%
Moreover, we split the remaining images into two disjoint sets at random---a training set and a test set---and filter out data from any expert who made less than $130$ and $20$ predictions in 
training and test set, respectively, and whose predicted labels in the training data do not cover all class labels.
%
After these preprocessing steps, the resulting training and test sets contain $1{,}257$ and $303$ images, respectively, annotated by $|\Hcal| = 114$ experts, where each image in the training and test set is annotated by at least two experts.
\begin{table}
    \centering
    \small
    \setlength{\tabcolsep}{4pt}
    \caption{Overall test accuracy}\label{tab:acc_table}
    \begin{tabular}{rccc}
      \toprule 
	    \bfseries Model & \bfseries $h, h' \in \Hcal$ & \bfseries $h, h' \in \psi$ & \bfseries $h \in \psi$, $h \in \psi'$ \\
      \midrule 
	    Gumbel-Max SI-SCM &  66.8\% & 79.9\% & 45.1\%\\
	    GNB & 48.9\% & 51.3\% & 45.1\%\\
	    GNB + CNB & 62.0\% & 66.0\% &55.2\%\\
      \bottomrule 
    \end{tabular}
    \vspace{-3mm}
\end{table}
\begin{figure*}[ht!]
        \centering
	\subfloat[][Gumbel-Max SI-SCM vs. GNB]{
        \stackunder{\includegraphics[width=0.4\textwidth]{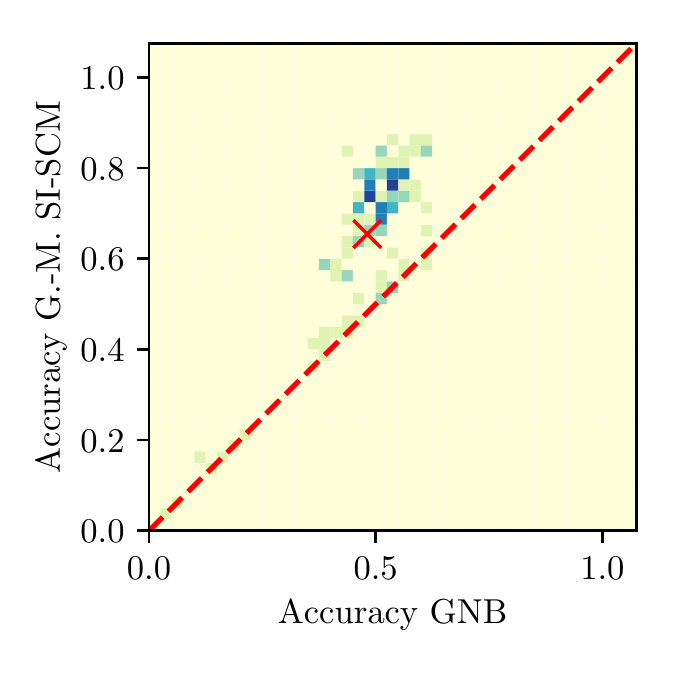}}{ $h, h' \in \Hcal$}
        \stackunder{\includegraphics[width=0.48\textwidth]{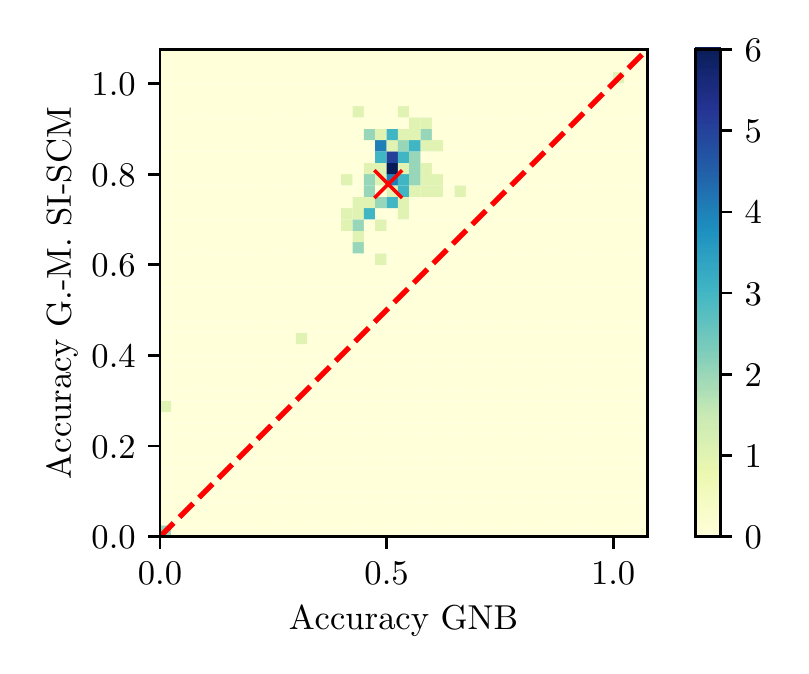}}{ $h, h' \in \psi$}
        }
	\\
	\subfloat[][Gumbel-Max SI-SCM vs. GNB+CNB]{
        \stackunder{\includegraphics[width=0.4\textwidth]{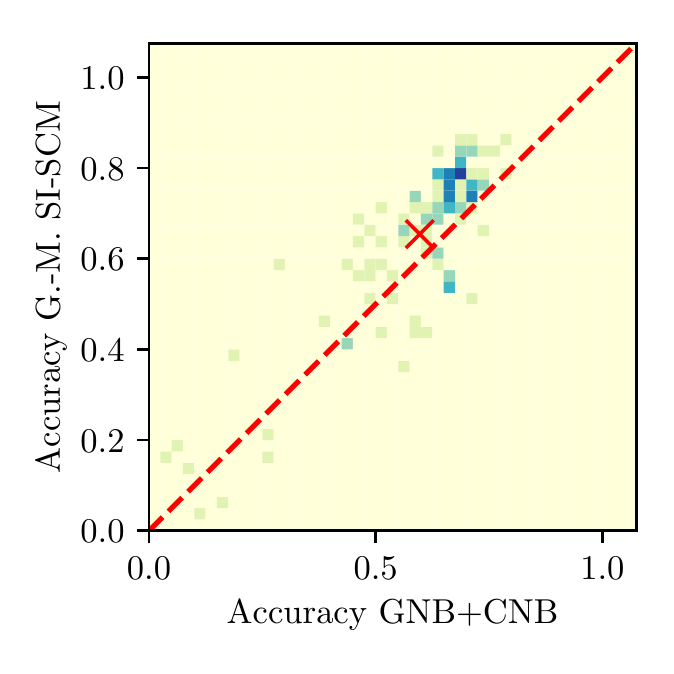}}{ $h, h' \in \Hcal$}
        \stackunder{\includegraphics[width=0.48\textwidth]{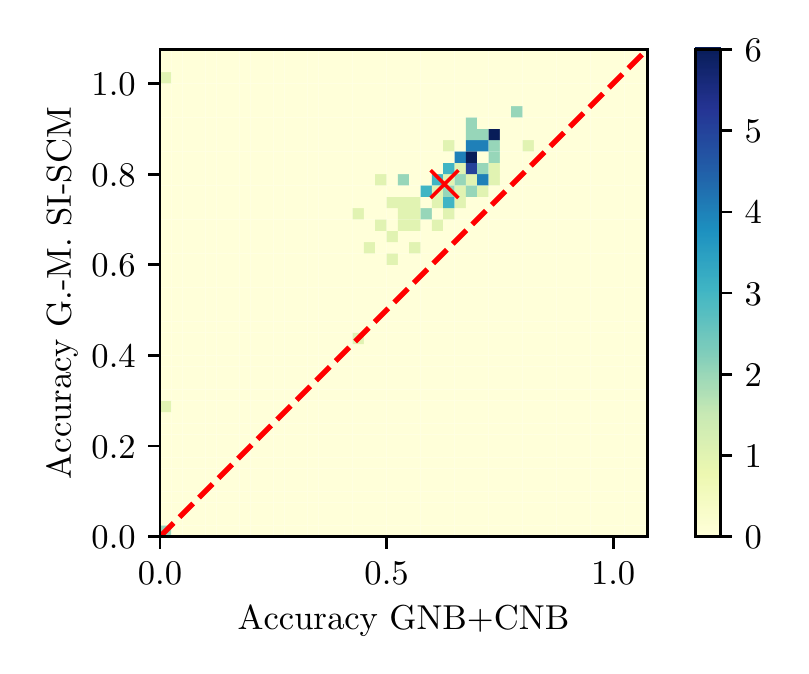}}{ $h, h' \in \psi$}
        }
        \vspace{-2mm}
        \caption{Per-expert test accuracy achieved by our model and both baselines on the CIFAR-10H dataset. In each panel, the $y$-axis always measures the per-expert test accuracy achieved by
        our method and the $x$-axis measures the per-expert accuracy achieved by one of the baselines.
        For each cell, the darkness is proportional to the number of experts with the corresponding test accuracies.}
        \vspace{-2mm}
\label{fig:results_real}
\end{figure*}
To find the groups of mutually similar experts underpinning our Gumbel-Max SI-SCM, we run Algorithm~\ref{alg:greedyalg} on the trai\-ning set.  
Within the Gumbel-Max SI-SCM, we estimate the conditional 
distribution $P^{\M \,;\, \text{do}[Z=\{h\}]}(Y_h \given X)$ for each expert $h$ using a Gaussian Naive Bayes model (GNB) trained using also the training set (one GNB per expert)\footnote{In 
the CIFAR-10H dataset, experts are assigned to images (presumably) at random. Therefore, it holds that $P^{\M \,;\, \text{do}[Z=\{h\}]}(Y_h \given X) = P(Y_h \given X, Z=\{h\})$ 
and we can use observational data to estimate the interventional conditional distribution $P^{\M \,;\, \text{do}[Z=\{h\}]}(Y_h \given X)$.}. 
%
%
%
Each GNB model uses $20$ dimensional feature vectors computed by running PCA on a $512$ dimensional normalized feature vector extracted using VGG19~\citep{simonyan2014very}.
Both during training and test, given an observed label prediction $Y_h$ by an expert $h$, we infer the prediction $Y_{h'}$ by another expert $h'$ using the most likely label under (an estimate of) the corresponding counterfactual distribution.
%
%
%
To estimate each counterfactual distribution, we use $T=1{,}000$ samples from the noise posterior distribution.
%

%

\xhdr{Baselines and evaluation metrics}
We compare the performance of our trained Gumbel-Max SI-SCM with two baselines (see also Figure~{\ref{fig:flowdiag_experiments}} in Appendix~{\ref{app:real}}): 

\noindent \hspace{0mm} --- The ``\emph{GNB}'' baseline uses only the same Gaussian Naive Bayes models (GNB), one per expert, used by our trained Gumbel-Max SI-SCM. 
More specifically, given an observed label prediction $Y_h$ by an expert $h$, it infers the prediction $Y_{h'}$ by another expert $h'$ using the most likely label under the estimate of the 
conditional distribution $P^{\M \,;\, \text{do}[Z=\{h'\}]}(Y_{h'} \given X)$ given by the corres\-pon\-ding~GNB. 

\noindent \hspace{0mm} --- The ``\emph{GNB + CNB}'' baseline uses the same Gaussian\- Naive Bayes models (GNB), one per expert, used by our trained Gumbel-Max SI-SCM and
a Categorical Naive Bayes (CNB) model, one per expert, that estimates $P^{\M\,;\,\text{do}[Z = \{h, h'\}]}(Y_{h'} \given Y_{h})$.\footnote{The CNB uses a ``one-hot'' encoding of the observed prediction
$Y_h$ as a single $|\Hcal|$-dimensional feature where, for each dimension, it uses an additional label value to denote than an expert'{}s label prediction has not been observed.} More specifically, 
given an observed label prediction $Y_h$ by an expert $h$, it infers the prediction $Y_{h'}$ by another expert $h'$ using the most likely label under the product of distributions 
$P^{\M \,;\, \text{do}[Z=\{h'\}]}(Y_{h'} \given X) \times P^{\M \,;\, \text{do}[Z = \{h, h'\}]}(Y_{h'} \given Y_{h})$, as estimated by the corres\-pon\-ding GNB (first term) and CNB (second term).
%

%
%
%
To compare the performance of our trained Gumbel-Max SI-SCM and both baselines, for each sample in the test set, we pick each of the corresponding expert label predictions $Y_h$ as the observed 
prediction in turn and infer the value of the other predictions $Y_{h'}$.
Here, we compute the overall accuracy as well as the per-expert accuracy and distinguish among three scenarios: (i) $h, h' \in \Hcal$; (ii) $h, h' \in \psi$; and, (iii) $h \in \psi, h' \in \psi', \psi \neq \psi'$.
%
%

\xhdr{Results} 
We start by reporting that, during the training of our Gumbel-Max SI-SCM, Algorithm~\ref{alg:greedyalg} found $352$ violations of the conditional stability condition between pairs of
experts and partitioned the experts into fifteen disjoint groups of mutually similar experts, where seven of these groups were singletons. Refer to Appendix~\ref{app:real} for more 
details regarding the groups identified by Algorithm~\ref{alg:greedyalg}. 

Next, we report the overall accuracy achieved by our model and the baselines in Table~\ref{tab:acc_table}. 
We find that, in general ($h, h' \in \Hcal$), our model infers the expert predictions more accurately than both baselines
%
%
and this competitive advantage comes from instances in which the observed prediction is by an expert $h$ who belongs to the same group 
of mutually similar experts as the expert $h'$ whose prediction we infer ($h, h' \in \psi$).
In fact, the GNB+CNB baseline is more accurate whenever both experts $h$ and $h'$ do not belong to the same group ($h \in \psi$, $h' \in \psi'$, $\psi \neq \psi'$).
Moreover, we also find that the GNB+CNB baseline infers the expert predictions more accurately whenever both experts belong to the same group of mutually 
similar experts identified by Algorithm~\ref{alg:greedyalg}.
In Appendix~\ref{app:real}, we report the confusion matrix of the above counterfactual predictions.

%
%



\begin{figure*}[t]
        \centering
        \includegraphics[width=0.5\textwidth]{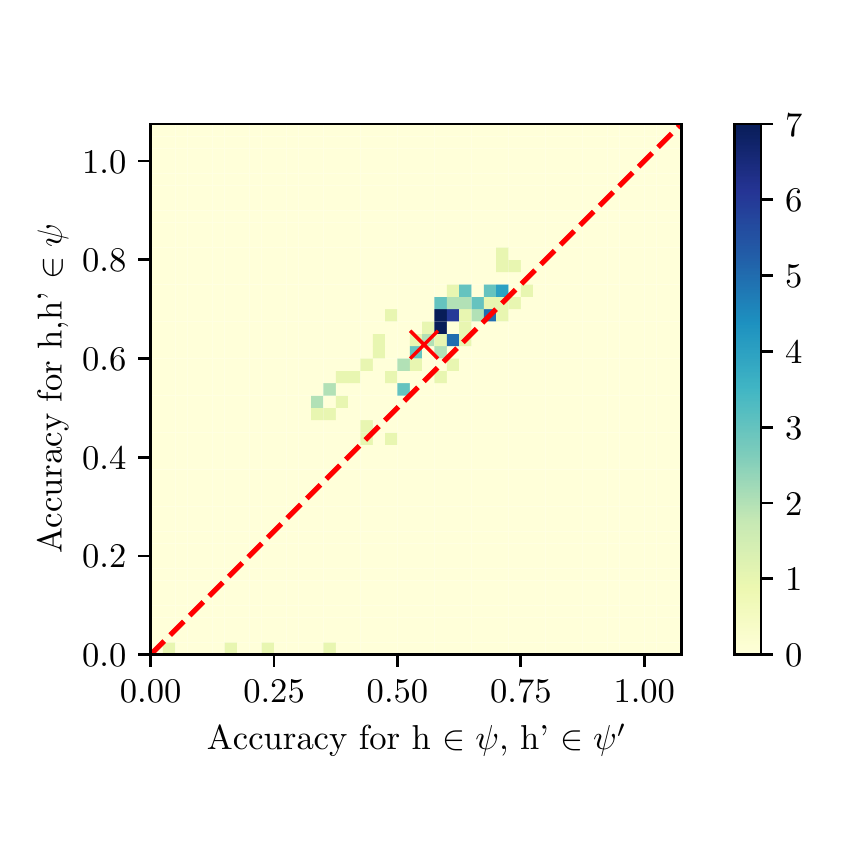}
        \vspace{-1mm}
        \caption{Per-expert test accuracy achieved by the baseline GNB+CNB on the preprocessed CIFAR-10H dataset. For each expert $h'$, the $y$-axis measures
        the test accuracy whenever the observed expert $h$ belongs to the same group of mutually similar experts as $h'$ and the $x$-axis measures
        the test accuracy whenever $h$ does not belong to the same group.
        For each cell, the darkness is proportional to the number of experts with the corresponding test accuracies.}
        \vspace{-2mm}
\label{fig:results_real_2}
\end{figure*}

Finally, we report the per-expert $h'$ accuracy achieved by our model and both baselines in Figure~\ref{fig:results_real}.\footnote{Whenever $h, h' \in \psi$, we could not compute the per-expert accuracy 
for $11$ experts---seven of these experts belong to singleton groups and the remaining four do not predict any of the same test samples predicted by other experts in their mutually similar groups.}
%
%
The results show that, in general ($h, h' \in \Hcal$), our model infers the expert predictions more accurately than the baselines for a majority 
of the experts ($103$ and $89$, out of $114$, compared to GNB and GNB+CNB, respectively). 
Moreover, if we restrict our attention to observed label predictions by experts $h$ belonging to the same group of mutually similar experts as the 
expert $h'$ whose prediction we infer ($h, h' \in \psi$), 
our model infers the expert prediction more accurately for almost all experts $h'$ ($100$ and $101$, out of $103$, compared to GNB and 
GNB+CNB, respectively). 
%
%
Additionally, Figure~\ref{fig:results_real_2} shows that, for most experts ($87$ out of $103$), the GNB+CNB baseline infers the expert predictions $Y_{h'}$ more 
accurately if the observed prediction $Y_{h}$ is by an expert $h$ belonging to the same group of mutually similar experts as the expert $h'$ ($h, h' \in \psi$) than if it is by an expert $h$ belonging to a different group ($h \in \psi$, $h' \in \psi'$, $\psi \neq \psi'$).


\vspace{-3mm}
\section{Conclusion}
\label{sec:conclusions}
\vspace{-2mm}
In this work, we have addressed the problem of inferring second opinions by experts from the perspective of counterfactual
inference.
We have focused on a multiclass classification setting and showed that, if experts make predictions on their own, the underlying
causal mechanism generating their predictions needs to satisfy a desirable set invariant property. 
Moreover, we have introduced the set invariant Gumbel-Max structural causal model, a new class of structural causal model 
whose structure and counterfactual predictions about second opinions by experts can be validated using interventional data.
%
%

Our work opens up many interesting avenues for future work. 
For example, we assume experts do not communicate before forming their opinion. 
Although this assumption may be satisfied in some real-world applications, it would be interesting to relax it.
Moreover, we have validated our model using a single real dataset. 
It would be valuable to validate 
our model using additional datasets 
from other applications. 
Finally, it would be important to carry out user studies in which the inferred second opinions provided by our model are shared
with domain experts (\eg, medical doctors). 

\vspace{2mm}

\xhdr{Acknowledgements} Gomez-Rodriguez acknowledges support from the European Research Council (ERC) under the European Union'{}s Horizon 2020 research and innovation programme (grant agreement No. 945719).

{
\bibliographystyle{unsrt}
\bibliography{cfact_second_opinions}

\begin{thebibliography}{27}
\providecommand{\natexlab}[1]{#1}
\providecommand{\url}[1]{\texttt{#1}}
\expandafter\ifx\csname urlstyle\endcsname\relax
  \providecommand{\doi}[1]{doi: #1}\else
  \providecommand{\doi}{doi: \begingroup \urlstyle{rm}\Url}\fi

\bibitem[Althabe et~al.(2004)Althabe, Beliz{\'a}n, Villar, Alexander, Bergel,
  Ramos, Romero, Donner, Lindmark, Langer, et~al.]{althabe2004mandatory}
Fernando Althabe, Jos{\'e}~M Beliz{\'a}n, Jos{\'e} Villar, Sophie Alexander,
  Eduardo Bergel, Silvina Ramos, Mariana Romero, Allan Donner, Gunilla
  Lindmark, Ana Langer, et~al.
\newblock Mandatory second opinion to reduce rates of unnecessary caesarean
  sections in latin america: a cluster randomised controlled trial.
\newblock \emph{The Lancet}, 363\penalty0 (9425):\penalty0 1934--1940, 2004.

\bibitem[Bica et~al.(2020)Bica, Jarrett, H{\"u}y{\"u}k, and van~der
  Schaar]{bica2020learning}
Ioana Bica, Daniel Jarrett, Alihan H{\"u}y{\"u}k, and Mihaela van~der Schaar.
\newblock Learning" what-if" explanations for sequential decision-making.
\newblock \emph{arXiv preprint arXiv:2007.13531}, 2020.

\bibitem[Brimberg et~al.(2017)Brimberg, Jani{\'c}ijevi{\'c}, Mladenovi{\'c},
  and Uro{\v{s}}evi{\'c}]{brimberg2017solving}
Jack Brimberg, Stefana Jani{\'c}ijevi{\'c}, Nenad Mladenovi{\'c}, and Dragan
  Uro{\v{s}}evi{\'c}.
\newblock Solving the clique partitioning problem as a maximally diverse
  grouping problem.
\newblock \emph{Optimization Letters}, 11\penalty0 (6):\penalty0 1123--1135,
  2017.

\bibitem[Burger et~al.(2020)Burger, Westerink, and Vrijsen]{burger2020outcomes}
Pascal~M Burger, Jan Westerink, and Bram~EL Vrijsen.
\newblock Outcomes of second opinions in general internal medicine.
\newblock \emph{PloS one}, 15\penalty0 (7):\penalty0 e0236048, 2020.

\bibitem[Dawid and Skene(1979)]{dawid1979maximum}
Alexander~Philip Dawid and Allan~M Skene.
\newblock Maximum likelihood estimation of observer error-rates using the em
  algorithm.
\newblock \emph{Journal of the Royal Statistical Society: Series C (Applied
  Statistics)}, 28\penalty0 (1):\penalty0 20--28, 1979.

\bibitem[Elmore et~al.(2015)Elmore, Longton, Carney, Geller, Onega, Tosteson,
  Nelson, Pepe, Allison, Schnitt, et~al.]{elmore2015diagnostic}
Joann~G Elmore, Gary~M Longton, Patricia~A Carney, Berta~M Geller, Tracy Onega,
  Anna~NA Tosteson, Heidi~D Nelson, Margaret~S Pepe, Kimberly~H Allison,
  Stuart~J Schnitt, et~al.
\newblock Diagnostic concordance among pathologists interpreting breast biopsy
  specimens.
\newblock \emph{Jama}, 313\penalty0 (11):\penalty0 1122--1132, 2015.

\bibitem[Gr{\"o}tschel and Wakabayashi(1989)]{grotschel1989cutting}
Martin Gr{\"o}tschel and Yoshiko Wakabayashi.
\newblock A cutting plane algorithm for a clustering problem.
\newblock \emph{Mathematical Programming}, 45\penalty0 (1):\penalty0 59--96,
  1989.

\bibitem[Gr{\"o}tschel and Wakabayashi(1990)]{grotschel1990facets}
Martin Gr{\"o}tschel and Yoshiko Wakabayashi.
\newblock Facets of the clique partitioning polytope.
\newblock \emph{Mathematical Programming}, 47\penalty0 (1):\penalty0 367--387,
  1990.

\bibitem[Guan et~al.(2018)Guan, Gulshan, Dai, and Hinton]{guan2018said}
Melody Guan, Varun Gulshan, Andrew Dai, and Geoffrey Hinton.
\newblock Who said what: Modeling individual labelers improves classification.
\newblock In \emph{Proceedings of the AAAI Conference on Artificial
  Intelligence}, volume~32, 2018.

\bibitem[Imbens and Rubin(2015)]{imbens2015causal}
Guido~W Imbens and Donald~B Rubin.
\newblock \emph{Causal inference in statistics, social, and biomedical
  sciences}.
\newblock Cambridge University Press, 2015.

\bibitem[Kerrigan et~al.(2021)Kerrigan, Smyth, and
  Steyvers]{kerrigan2021combining}
Gavin Kerrigan, Padhraic Smyth, and Mark Steyvers.
\newblock Combining human predictions with model probabilities via confusion
  matrices and calibration.
\newblock \emph{Advances in Neural Information Processing Systems}, 34, 2021.

\bibitem[Krizhevsky et~al.(2009)Krizhevsky, Hinton,
  et~al.]{krizhevsky2009learning}
Alex Krizhevsky, Geoffrey Hinton, et~al.
\newblock Learning multiple layers of features from tiny images.
\newblock 2009.

\bibitem[Leape(1989)]{leape1989unnecessary}
Lucian~L Leape.
\newblock Unnecessary surgery.
\newblock \emph{Health Services Research}, 24\penalty0 (3):\penalty0 351, 1989.

\bibitem[Lim et~al.(2021)Lim, Ji, Oberst, Blecker, Horwitz, and
  Sontag]{lim2021finding}
Justin Lim, Christina Ji, Michael Oberst, Saul Blecker, Leora Horwitz, and
  David Sontag.
\newblock Finding regions of heterogeneity in decision-making via expected
  conditional covariance.
\newblock \emph{Advances in Neural Information Processing Systems}, 34, 2021.

\bibitem[Lorberbom et~al.(2021)Lorberbom, Johnson, Maddison, Tarlow, and
  Hazan]{lorberbom2021learning}
Guy Lorberbom, Daniel Johnson, Chris~J Maddison, Daniel Tarlow, and Tamir
  Hazan.
\newblock Learning generalized gumbel-max causal mechanisms.
\newblock \emph{Advances in Neural Information Processing Systems}, 34, 2021.

\bibitem[Maddison et~al.(2015)Maddison, Tarlow, and Minka]{maddison2015a}
Chris~J. Maddison, Daniel Tarlow, and Tom Minka.
\newblock A* sampling, 2015.

\bibitem[Noorbakhsh and Rodriguez(2021)]{noorbakhsh2021counterfactual}
Kimia Noorbakhsh and Manuel~Gomez Rodriguez.
\newblock Counterfactual temporal point processes.
\newblock \emph{arXiv preprint arXiv:2111.07603}, 2021.

\bibitem[Oberst and Sontag(2019)]{oberst2019}
Michael Oberst and David Sontag.
\newblock Counterfactual off-policy evaluation with gumbel-max structural
  causal models.
\newblock In \emph{International Conference on Machine Learning}, pages
  4881--4890. PMLR, 2019.

\bibitem[Pearl(2009)]{pearl2009causality}
Judea Pearl.
\newblock \emph{Causality}.
\newblock Cambridge university press, 2009.

\bibitem[Peterson et~al.(2019)Peterson, Battleday, Griffiths, and
  Russakovsky]{peterson2019human}
Joshua~C Peterson, Ruairidh~M Battleday, Thomas~L Griffiths, and Olga
  Russakovsky.
\newblock Human uncertainty makes classification more robust.
\newblock In \emph{Proceedings of the IEEE/CVF International Conference on
  Computer Vision}, pages 9617--9626, 2019.

\bibitem[Raghu et~al.(2019)Raghu, Blumer, Sayres, Obermeyer, Kleinberg,
  Mullainathan, and Kleinberg]{raghu2019direct}
Maithra Raghu, Katy Blumer, Rory Sayres, Ziad Obermeyer, Bobby Kleinberg,
  Sendhil Mullainathan, and Jon Kleinberg.
\newblock Direct uncertainty prediction for medical second opinions.
\newblock In \emph{International Conference on Machine Learning}, pages
  5281--5290. PMLR, 2019.

\bibitem[Simonyan and Zisserman(2014)]{simonyan2014very}
Karen Simonyan and Andrew Zisserman.
\newblock Very deep convolutional networks for large-scale image recognition.
\newblock \emph{arXiv preprint arXiv:1409.1556}, 2014.

\bibitem[Straitouri et~al.(2022)Straitouri, Wang, Okati, and
  Rodriguez]{straitouri2022provably}
Eleni Straitouri, Lequn Wang, Nastaran Okati, and Manuel~Gomez Rodriguez.
\newblock Provably improving expert predictions with conformal prediction.
\newblock \emph{arXiv preprint arXiv:2201.12006}, 2022.

\bibitem[Tsirtsis et~al.(2021)Tsirtsis, De, and
  Rodriguez]{tsirtsis2021counterfactual}
Stratis Tsirtsis, Abir De, and Manuel Rodriguez.
\newblock Counterfactual explanations in sequential decision making under
  uncertainty.
\newblock \emph{Advances in Neural Information Processing Systems}, 34, 2021.

\bibitem[Van~Such et~al.(2017)Van~Such, Lohr, Beckman, and
  Naessens]{van2017extent}
Monica Van~Such, Robert Lohr, Thomas Beckman, and James~M Naessens.
\newblock Extent of diagnostic agreement among medical referrals.
\newblock \emph{Journal of evaluation in clinical practice}, 23\penalty0
  (4):\penalty0 870--874, 2017.

\bibitem[Welinder and Perona(2010)]{welinder2010online}
Peter Welinder and Pietro Perona.
\newblock Online crowdsourcing: rating annotators and obtaining cost-effective
  labels.
\newblock In \emph{2010 IEEE Computer Society Conference on Computer Vision and
  Pattern Recognition-Workshops}, pages 25--32. IEEE, 2010.

\bibitem[Zhang et~al.(2016)Zhang, Wu, and Sheng]{zhang2016learning}
Jing Zhang, Xindong Wu, and Victor~S Sheng.
\newblock Learning from crowdsourced labeled data: a survey.
\newblock \emph{Artificial Intelligence Review}, 46\penalty0 (4):\penalty0
  543--576, 2016.

\end{thebibliography}
}

\newpage
\appendix
\onecolumn

%

\section{Proofs}
\label{app:awesomeproofs}
\xhdr{Proof of Theorem~\ref{thm:SCMsetinvariant}}
Let $\zeta \subseteq \zeta' \subseteq \Hu$ and both non-empty. Then, for any $x \in \X, u \in \U$, we have that
	\begin{equation*}
	(f(x,\zeta',u))_\zeta = ((f_h(x,u))_{h\in \zeta'})_{\zeta} = \ (f_h(x,u))_{h\in \zeta} = f(x,\zeta,u).
	\end{equation*}

\xhdr{Proof of Theorem~\ref{thm:SCMequivalence}} 
%
%
	Let $\M'$ be constructed from $\M$ by changing causal mechanism $f$ with $f'$.
	To prove equivalence of $\M$ and $\M'$, we only need to show that, for any $u\in \U, x \in \X, \zeta \in \mathcal{P}(\Hu)\setminus \emptyset$, it holds that
	\begin{equation}
		\vecy = f(x,\zeta,u) \Longleftrightarrow \vecy = f'(x,\zeta,u),
	\end{equation}
as only the causal mechanism was altered in the construction of $\M'$.
Let $h \in \zeta$ be an arbitrary expert, then
	\begin{equation*} 
		(f'(x,\zeta,u))_h \overset{def.}{=} f(x,\{h\},u) = (f(x,\zeta,u))_h,
	\end{equation*}
where the last equality holds because $f$ is set invariant. 
Thus, $f(x,\zeta,u)=f'(x,\zeta,u)$ for all $x \in \X, \zeta \in \Hu, u \in \U$.

\xhdr{Proof of Theorem~\ref{thm:equalprop}}
%
%
For clarity, we explicitly write $\vecY_{Z=\{h'\}}$ and $\vecY_{Z=\zeta}$ to better distinguish the two interventional outcomes. 
	For discrete probability distribution $P(U)$ the right probability is given by 
	\begin{align*}
		& P^{M;do[Z:=\zeta']}(\vecY_\zeta=\vecy \mid X=x) 
		= \sum_{u \in \U} P(U=u) \cdot \mathbbm{1}{[f(x,\zeta',u))_\zeta=\vecy)]},
	\end{align*}
	whereas the left is given by
	\begin{align*}
		 P^{M;do[Z:=\zeta]}(\vecY=\vecy \mid X=x) 
		= \sum_{u \in \U} P(U=u) \cdot \mathbbm{1}{[f(x,\zeta,u)=\vecy]}.
	\end{align*}
	Because $f$ is a set invariant mechanism over $Z$, $(f(x,\zeta',u))_\zeta=f(x,\zeta,u)$, thus,
	\begin{align*}
		\sum_{u \in \U} P(U=u) \cdot \mathbbm{1}{[f(x,\zeta,u)=\vecy]}
		= & \sum_{u \in \U} P(U=u) \cdot \mathbbm{1}{[f(x,\zeta',u))_\zeta=\vecy]}.
	\end{align*}
	The proof is analogous for the continuous probability distributions $P(U)$. 

\xhdr{Proof of Corollary~\ref{cor:equalprop_h}}
	Choose $\zeta=\{h\}$ in Theorem~\ref{thm:equalprop} and note that abusing notation $\vecY_{Z=\zeta}$ is in this case equivalent to $Y_h$.

\xhdr{Proof of Corollary~\ref{cor:equalcounterfactuals}}
	Using the definition of counterfactual distributions, the proof is analogous to the proof of Theorem~\ref{thm:equalprop} but using the posterior distribution $P(U\mid X=x, Z=\{h\},Y_h=c)$. 
	Let $\zeta \subseteq H$ be so that $h, h' \in \zeta$.
For all $c \in \Y$, by definition, we have that
        \begin{align}
		P^{M\mid X=x, Z=\{h\},\vecY=c ;do[Z:=\{h'\}]}(\vecY=c')
		 = 
		  \sum_{u \in \U} P(U =u \mid X, (\vecY_{Z=\{h\}})_{h}=c) \cdot \mathbbm{1}{[f(x,\{h'\},u)=c']}. \label{eq:cprop}
        \end{align}
	Using that $\M$ is set invariant we get that
        \begin{align*}
		(\vecY_{Z=\{h\}})_{h}=f(x,\{h\} ,U) = (f(x,\zeta,U))_{h} = (\vecY_{Z=\zeta})_{h}
		\quad \text{ and } \quad
		\mathbbm{1}{[f(x,\{h'\},u))=c']} = \mathbbm{1}{[(f(x,\zeta,u))_{h'}=c']}.
        \end{align*}
	Thus, Eq.~\eqref{eq:cprop} is equal to
        \begin{align*}
		\sum_{u \in \U} P(U =u \mid X, (\vecY_{Z=\zeta})_{h}=c) \cdot \mathbbm{1}{[(f(x,\zeta,u))_{h'}=c']}
		\overset{def.}{=} P^{M;do[Z:=\zeta]}(Y_{h'}=c'\mid X=x, Y_{h}=c).
        \end{align*}
%
%

\xhdr{Proof of Theorem~\ref{thm:pcs_equivalence}}
%
        Note that for all $\zeta_1, \zeta_2,\zeta_3 \subseteq H$ so that $h \in \zeta_1, h' \in \zeta_2, h,h'\in \zeta_3$ holds that
        \begin{align}
                P^{\M\mid X=x, Z=\zeta_1, Y_{h}=c;do[Z=\zeta_2]}(Y_{h'}=c')=0 &\iff P^{\M\mid X=x, Z=\{h\}, Y_{h}=c;do[Z=\{h'\}]}(Y_{h'}=c')=0,
                \label{eq:equivalence1}
                \\ 
                P^{\M\mid X=x, Z=\{h\}, Y_{h}=c;do[Z=\{h'\}]}(Y_{h'}=c')=0 &\iff P^{\M;do[Z=\zeta_3]}(Y_{h'}=c'\mid X =x, Z=\zeta_3, Y_{h}=c)=0,
                \label{eq:equivalence2}
        \end{align}
        where Eq.~\eqref{eq:equivalence1} follows from the definition of set invariance and Eq.~\eqref{eq:equivalence2} follows from Corollary~\ref{cor:equalcounterfactuals}.
        Recall that $p_\zeta(h,c) := P^{\M;do[Z=\zeta]}(Y_{h}=c\mid X)$ and $p_h(c) := P^{\M;do[Z=\{h\}]}(Y_{h}=c\mid X)$.

	It follows from Corollary~\ref{cor:equalprop_h} that, for all $h \in H$, $c \in \Y$ and $\zeta,\zeta'$ so that $h \in \zeta$ and $h \in \zeta'$, we have that
        \begin{align*}
		p_{\zeta}(h,c) = p_h(h,c)= p_{\zeta'}(h,c).
        \end{align*}
	Thus, following implications hold
        \begin{align*}
		 \frac{p_{\zeta_2}(h',c)}{p_{\zeta_1}(h,c)} \geq \frac{p_{\zeta_2}(h', c')}{p_{\zeta_1}(h,c')}
		  \iff   
		 \frac{p_{h'}(h',c)}{p_{h}(h,c)} \geq \frac{p_{h'}(h',c')}{p_{h}(h,c')}
		 \iff  
		 \frac{p_{\zeta_3}(h',c)}{p_{\zeta_3}(h,c)} \geq \frac{p_{\zeta_3}(h', c')}{p_{\zeta_3}(h,c')}.
        \end{align*}
        With these set of implications, it is straight forward to imply one statement from the other.



\xhdr{Proof of Theorem~\ref{thm:gumbel}}
Let $\psi$ be a subgroup in $\Psi$ so that $|\psi|\geq 2$. Let $h$ and $h'$ denote two arbitrary experts in subgroup $\psi$.
        As the Gumbel-Max SI-SCM $\M(\Psi)$ is set invariant, it is enough to show that pairwise conditional stability condition is satisfied for pair $h$ and $h'$.
        Analogously to \citet{oberst2019}, we proceed by proving the contrapositive, that for all sets $\zeta$, so that $h,h' \in \zeta$, and $c\neq c'$
        \begin{align*}
                 P^{\M(\Psi);do[Z=\zeta]}(Y_{h'}=c' \mid X, Y_h=c) \neq 0
                 \implies \frac{p_\zeta(h',c)}{p_\zeta(h,c)}< \frac{p_\zeta(h',c')}{p_\zeta(h,c')}.
        \end{align*}
        If the conditional probability is positive, almost surely there must exist Gumbel noise variables $g_{\psi,c}$ and $g_{\psi,c'}$ such that
        \begin{align*}
                \log P(Y_h=c \mid X) + g_{\psi,c} &> \log P(Y_h=c' \mid X) + g_{\psi,c'} \\ 
                \log P(Y_{h'}=c \mid X) + g_{\psi,c} &< \log P(Y_{h'}=c' \mid X) + g_{\psi,c'},
        \end{align*}
        as the submechanisms $f_h$ and $f_{h'}$ of each expert share the noise vector of the subgroup $\psi$.

        Recall that, by set invariance, $p_\zeta(h,c)=P(Y_h=c \mid X)$ for all $\zeta,h,c$. Hence, we can substitute the probabilities in both inequalities.
        Then, we further subtract the first inequality from the second which cancels out the Gumbel noises.
        Finally, using the properties of the logarithm function, the inequality is rearranged deriving the implication.
        \begin{align*}
                \log p_\zeta(h',c) - \log p_\zeta(h,c) &< \log p_\zeta(h',c') -\log p_\zeta(h,c'), \\
                \frac{ p_\zeta(h',c)}{ p_\zeta(h,c)} &< \frac{p_\zeta(h',c')}{p_\zeta(h,c')}.
        \end{align*}
        This proves that the Gumbel-Max SI-SCM $\M(\Psi)$ satisfies the pairwise conditional stability condition
        \begin{equation*}
                \frac{p_\zeta(h',c)}{p_\zeta(h,c)}\geq \frac{p_\zeta(h',c')}{p_\zeta(h,c')}
                \implies P^{\M(\Psi);do[Z=\zeta]}(Y_{h'}=c' \mid X, Y_h=c) = 0,
        \end{equation*}
        for any two experts in the same subgroup in $\Psi$.

\section{Randomized Greedy Algorithm for the Clique Partitioning Problem}
\label{sec:algorithm}
The idea behind the simple greedy randomized Algorithm \ref{alg:greedyalg} is to sequentially grow a clique starting from a random vertex in $G'$ until no vertices can be added, remove this clique from the graph and repeat this process on the remaining graph until no vertices are left.
For the current clique $\psi$ the set of vertices that can be added, called candidate set, consists of vertices that
have edges to all the vertices in $\psi$ so that the sum these edge weights is non-positive.
The expert with minimum sum is added next to the clique, and the candidate set is updated.

The updated candidate set is a subset of the previous set, thus, the sum of edge weights of a vertex connected to the updated clique can be computed in constant time by considering the previous value and the weight of the edge to the newly added vertex. 
If no edge exists, the vertex can be removed from the candidate set.
Algorithm \ref{alg:greedyalg} can thus be implemented in $O(|\mathcal{E}|)$.
\begin{algorithm}[h]
    \caption{Greedy Algorithm for the Clique Partitioning Problem,
	 $N(\psi)$ denotes the set of vertices not in $\psi$ with edges to all vertices in $\psi$ }
	\label{alg:greedyalg}
    \begin{algorithmic}
	    \STATE \textbf{Input:} weighted, undirected graph $G=(H,E,w)$
	\WHILE{$G$ not empty}
	   \STATE pick random vertex $h$
	   \STATE $\psi \gets h$
	   \WHILE{$N(\psi)$ not empty}
		\STATE $h^* \gets \arg \min_{h' \in N(\psi)} \sum_{h\in \psi} w(\{h',h\})$
		\IF{ $\sum_{h\in \psi} w(\{h^*,h\})\leq0$}
			\STATE $\psi \gets h^*$
		\ELSE	\STATE break
		\ENDIF
	    \ENDWHILE
	    \STATE $\Psi \gets \psi$
	    \STATE delete $\psi$ from $G$
	    \STATE $\psi = \emptyset$
	\ENDWHILE
	\STATE \textbf{return} $\Psi$
	\end{algorithmic}
\end{algorithm}

We note that, since the algorithm minimizes the sum of weights for each clique sequentially, it's performance in recovering a partition minimizing the overall sum depends on the sequence of sampled vertices which we start each clique from. To stabilize the algorithm's performance, one can rerun the algorithm a few times and choose among the returned partitions the one minimizing the overall sum of edge weights between vertices in the same set (see objective function in optimization problem~\ref{eq:optproblem_graph}). In the experiments on real (synthetic) data, we run Algorithm~\ref{alg:greedyalg} $10$ ($5$) times.

\section{Experiments on Synthetic Data}
\label{sec:synthetic}
In this section, we assess the performance of Algorithm~\ref{alg:greedyalg} at recovering the groups of mutually similar experts underpinning our Gumbel-Max SI-SCM
using synthetic data. 

\xhdr{Experimental setup}
We consider a synthetic prediction task with $k = 5$ labels and $20$ features per sample, whose values we sample uniformly 
at random from the interval $[0, 1]$, and a set $\Hcal$ of $48$ synthetic experts.
%
%
%
These synthetic experts make label predictions according to a Gumbel-Max SI-SCM with five disjoint groups of mutually
similar experts $\Psi$, \ie, each expert within a group $\psi \in \Psi$ use the same Gumbel noise within the model\footnote{The 
groups in the partition $\Psi$ contain $6$, $7$, $11$, $11$ and $13$ experts.}.
Moreover, for each expert, the probability $P(Y_h=c \mid X=x)$ is given by a multinomial logit model with random weight coefficients 
$w=(w_1, \dots, w_5)$, which we also sample uniformly at random from the interval $[0, 1]$ independently for each expert, \ie,
\begin{equation} \label{eq:multdistr}
	P(Y_h=c \mid X=x) = \frac{\exp(w_c\cdot x)}{\sum_{j \in \Y} \exp(w_j\cdot x)}. 
\end{equation}
%


We measure the performance of Algorithm~\ref{alg:greedyalg} at recovering the partition $\Psi$ given the true probabilities $P(Y_h=c \mid X=x)$ under different amounts 
of training data $m$ and sparsity levels $s \in (0,1)$.
The sparsity level $s$ controls the average number of observed expert predictions per sample, \ie, for each sample, all experts make a prediction but we only
observe $\max\{2, (1-s) |\Hcal|\}$, picked at random.
Here, note that, as the sparsity level $s$ decreases (increases) and the amount of training data increases (decreases), it is easier (harder) to recover the partition $\Psi$.
%

As a measure of the difficulty of each inference problem, we will use the edge ratio $r$, defined as the fraction of pair of experts who belong to the same group $\psi \in \Psi$, among all 
pairs whose predictions did not violate conditional stability and were at least once observed for the same sample.
As performance metrics, we will use:
\begin{itemize}
	\item The adjusted random index (ARI), which measures similarity between the partition $\hat{\Psi}$ returned by Algorithm~\ref{alg:greedyalg} and the true partition $\Psi$. 
	Its value lies in the interval $[0,1]$ where $1.0$ means full recovery and $0.0$ means a completely random partition was recovered with no similarity to the true one.

	\item The average $0/1$-loss on a held-out set (with $1000$ samples) of a predictor that, given an observed label $Y_h$, returns the most likely label $Y_{h'}$ under 
	$P^{\M(\hat{\Psi}) \given X=x, Z=\{h\},\vecY=y_{h}\,;\, \text{do}(Z=\{h'\})(\vecY)}$, \ie,
		the inferred counterfactual distribution, estimated using $500$ samples.
\end{itemize}
As a point of comparison, for the second performance metric, we also compute the average $0/1$-loss over the same held-out set of two other predictors that, given an observed label $Y_h$, return 
the most likely label $Y_{h'}$ under the true counterfactual distributions $P^{\M(\Psi) \given X=x, Z=\{h\}\vecY=y_{h}; \text{do}(Z=\{h'\})(\vecY)}$ and the counterfactual distribution $P^{\M(\Hcal) 
\given X=x, Z=\{h\}\vecY=y_{h}; \text{do}(Z=\{h'\})(\vecY)}$, respectively.
%
%
%
%
\begin{figure*}[t]
        \centering
        	\subfloat[Adjusted random index (ARI)]{\includegraphics[width=0.32\textwidth]{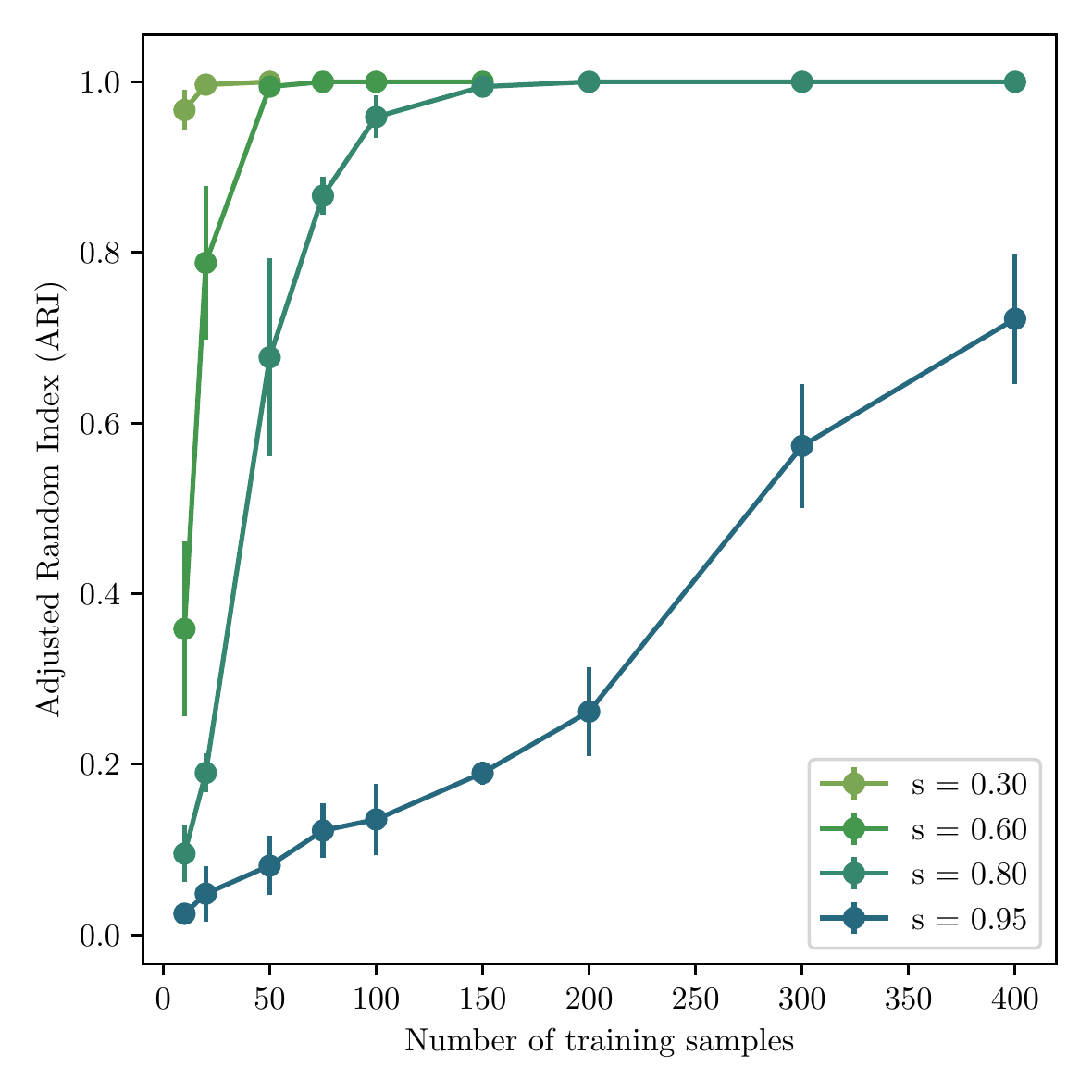}}
	\hspace{1mm}
        \subfloat[Average 0/1-loss on a held-out set]{\includegraphics[width=0.32\textwidth]{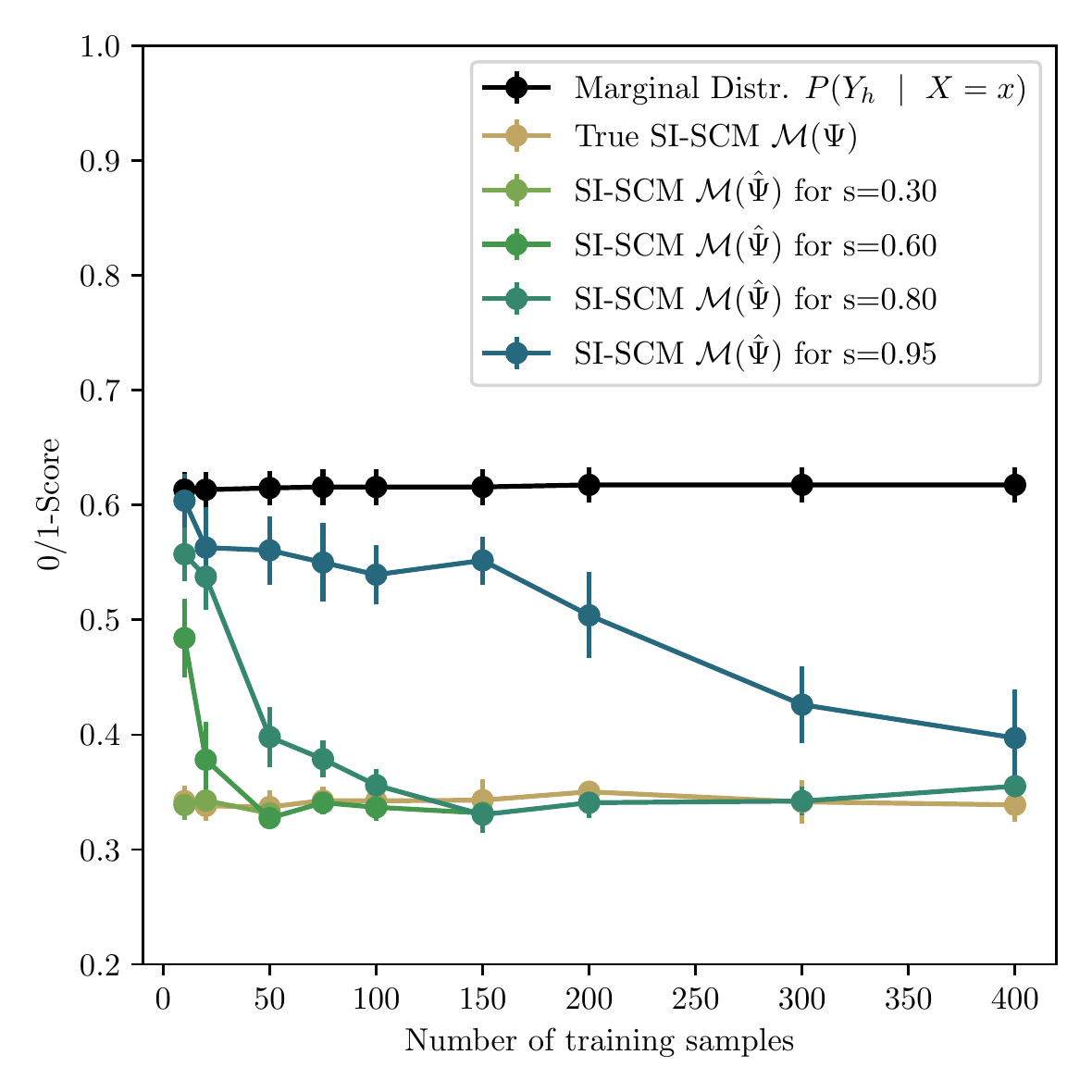}}
        \hspace{1mm}
	\subfloat[Edge ratio]{\includegraphics[width=0.32\textwidth]{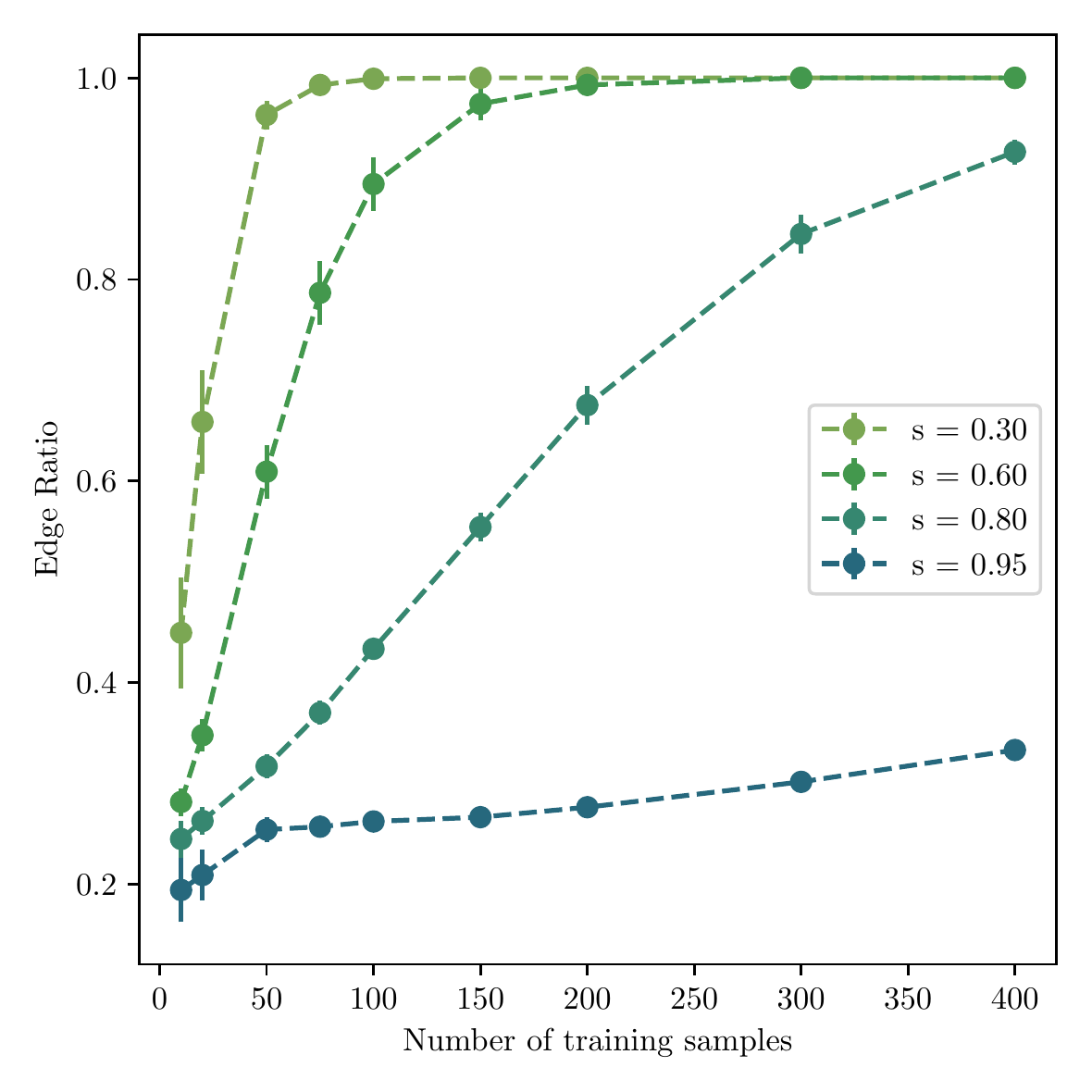}}
        \caption{Performance of Algorithm~\ref{alg:greedyalg} at recovering the partition $\Psi$ given the true probabilities $P(Y_h=c \mid X=x)$ under different amounts 
of data $m$ and sparsity levels $s \in (0,1)$. 
	To compute the mean and the standard deviation in panels (a), (b) and (c), we run each experiment five times.
	%
        }
\label{fig:results_synthetic}
\end{figure*}

\xhdr{Results}
%
%
%
Figure~\ref{fig:results_synthetic} summarizes the results, which show that, as long as the edge ratio $r > 0.3$, the inferred partition $\hat{\Psi}$ is 
very similar to the true partition $\Psi$ (\ie, the value of ARI is very close to $1$) and the $0/1$-losses of the predictors that use $\M(\hat{\Psi})$ and $\M(\Psi)$ respectively are very similar.
%
%
%
%
%
Here, note however that, even the predictor that uses the true model $\M(\Psi)$ has a non zero $0/1$-loss is not error free because, given an observed expert prediction $Y_h$ and feature vector $x$, 
the expert prediction $Y_{h'}$ is not deterministic.

\section{Additional Figures for Experiments on Real Data}
\label{app:real}
\begin{figure*}[!h]
        \centering
        \includegraphics[width=1.0\textwidth]{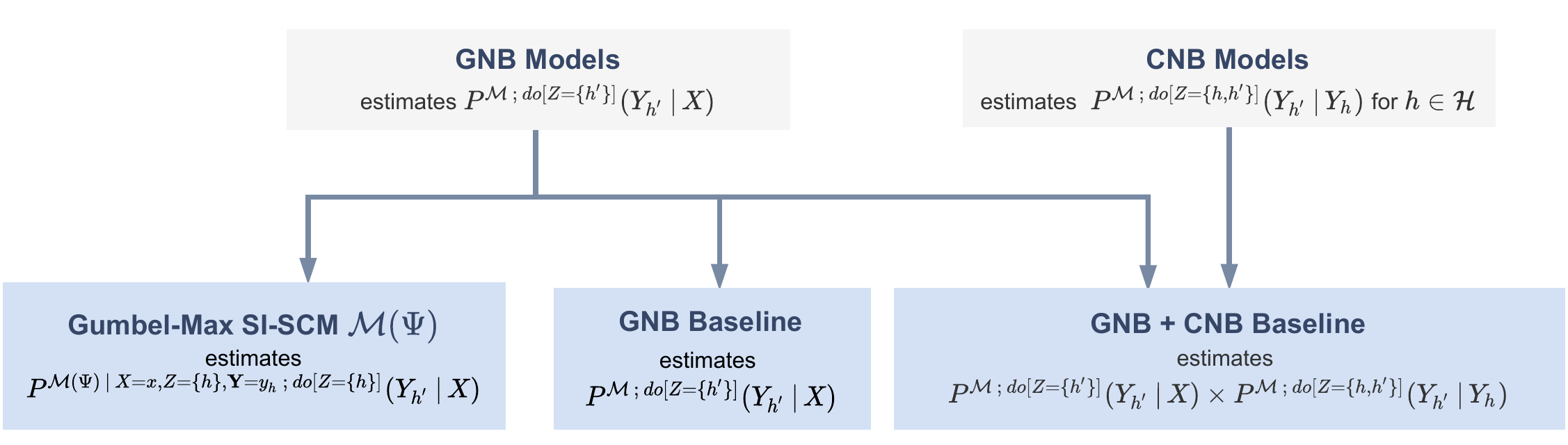}
        \vspace{-1mm} 
        \caption{Different models used in our experiments on real data.}
\label{fig:flowdiag_experiments}
\end{figure*}
\begin{figure*}[!h]
        \centering
        \includegraphics[width=0.7\textwidth]{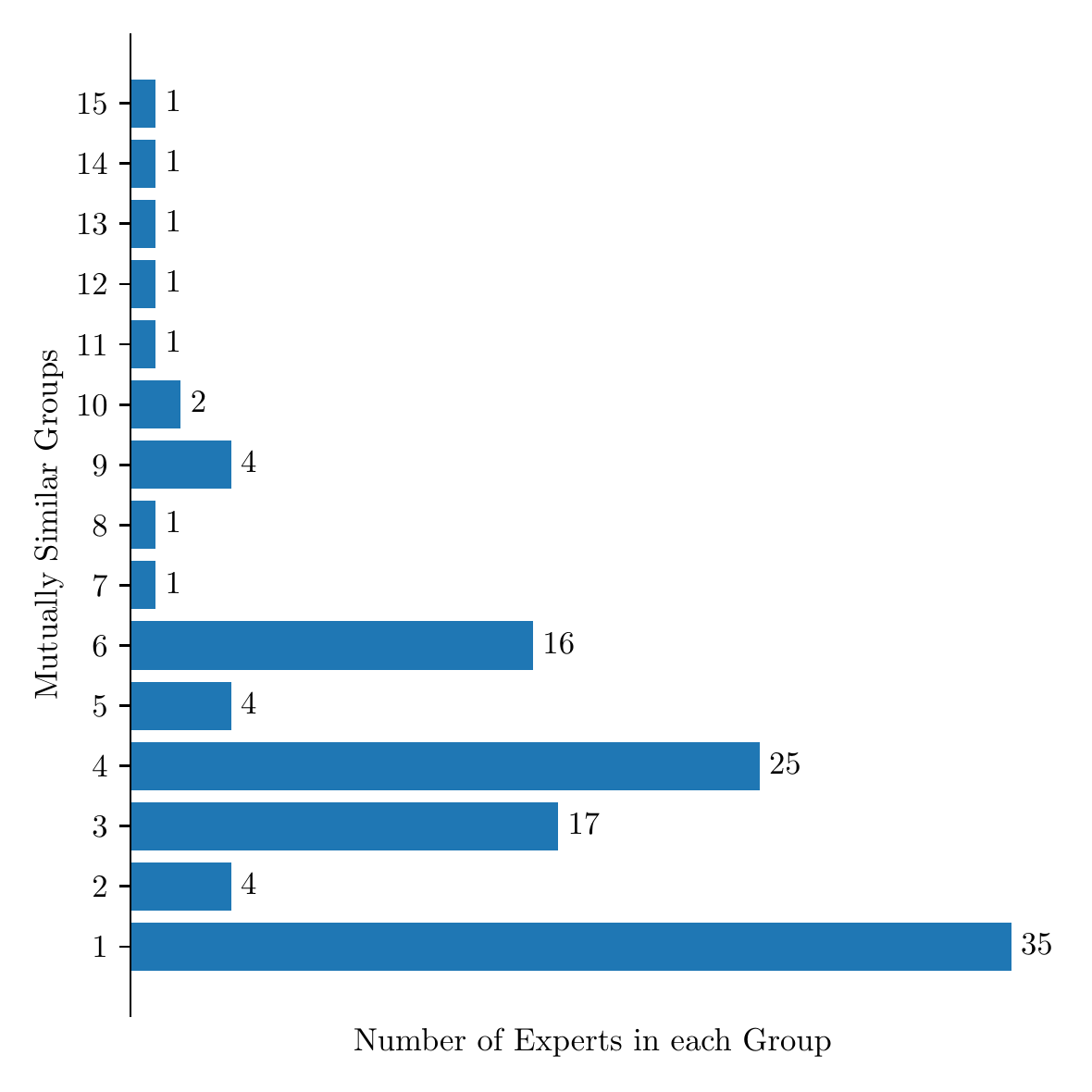}
        \vspace{-1mm}
	\caption{Size of the mutually similar expert groups returned by Algorithm~\ref{alg:greedyalg} for the preprocessed CIFAR-10H dataset.}
\label{fig:results_groups}
\end{figure*}
\begin{figure*}[!t]
        \centering
                \subfloat[\hspace{10mm} Gumbel-Max SI-SCM]{\includegraphics[width=0.5\textwidth]{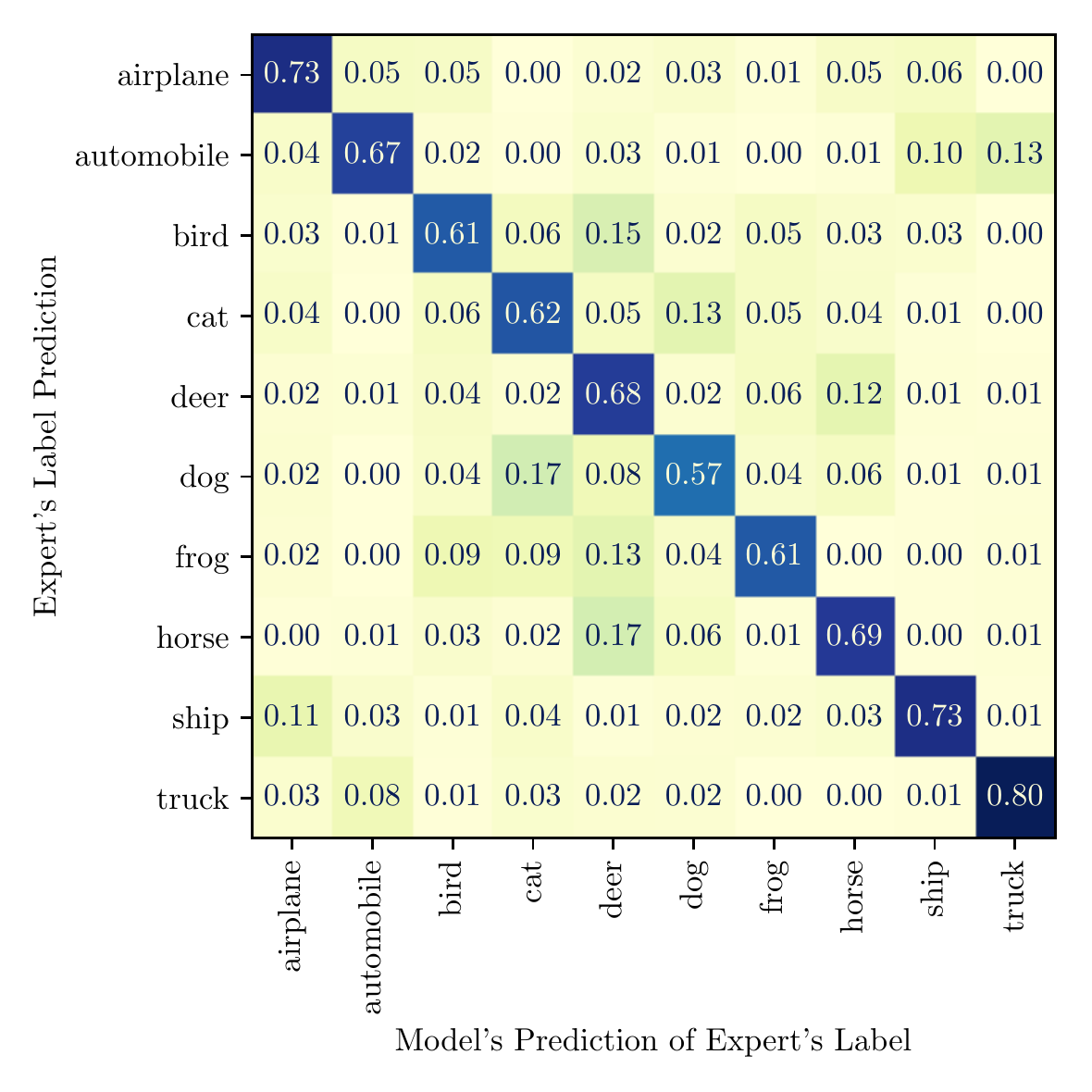}}
        \subfloat[\hspace{10mm}GNB]{\includegraphics[width=0.5\textwidth]{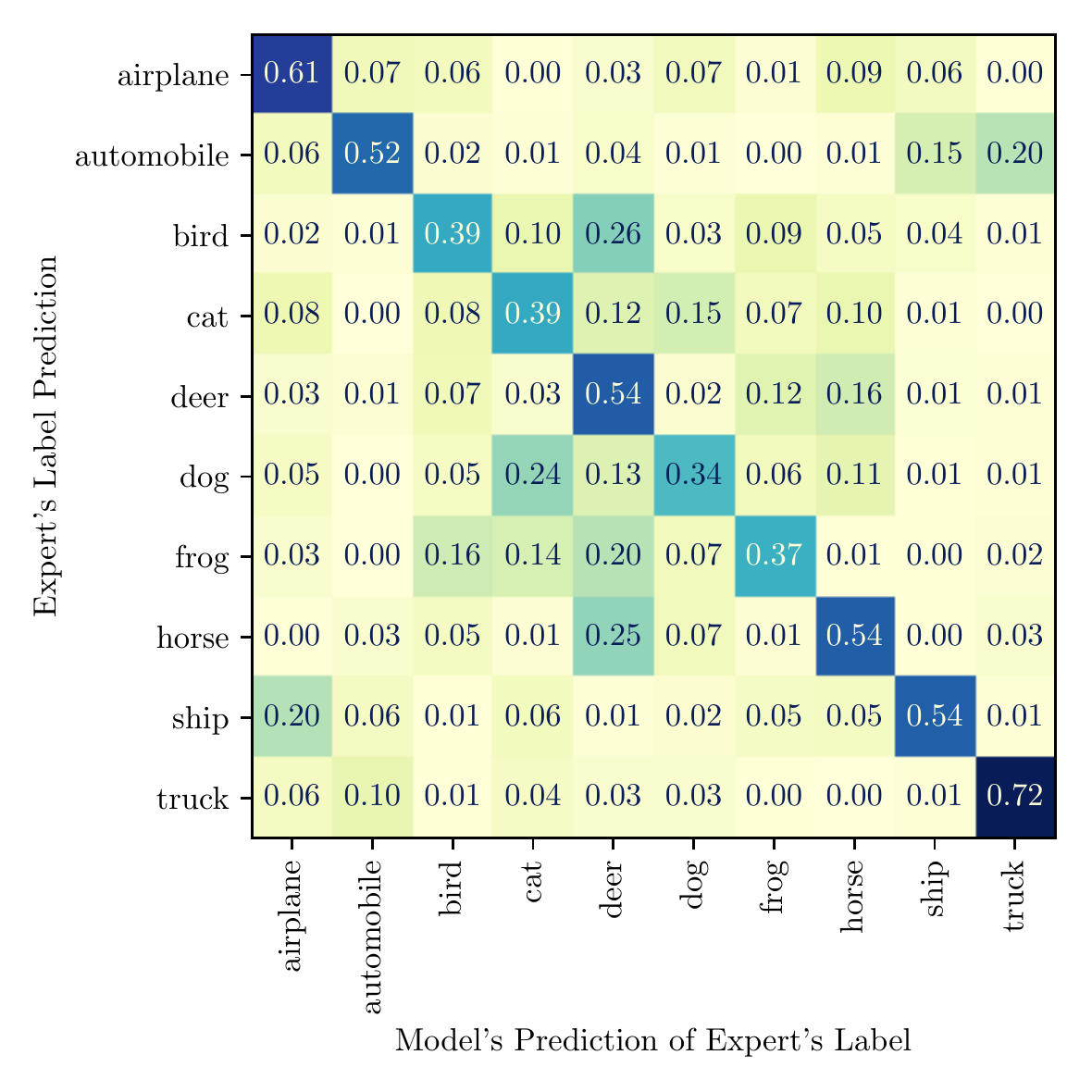}}
	\\
        \subfloat[\hspace{10mm} GNB+CNB]{\includegraphics[width=0.5\textwidth]{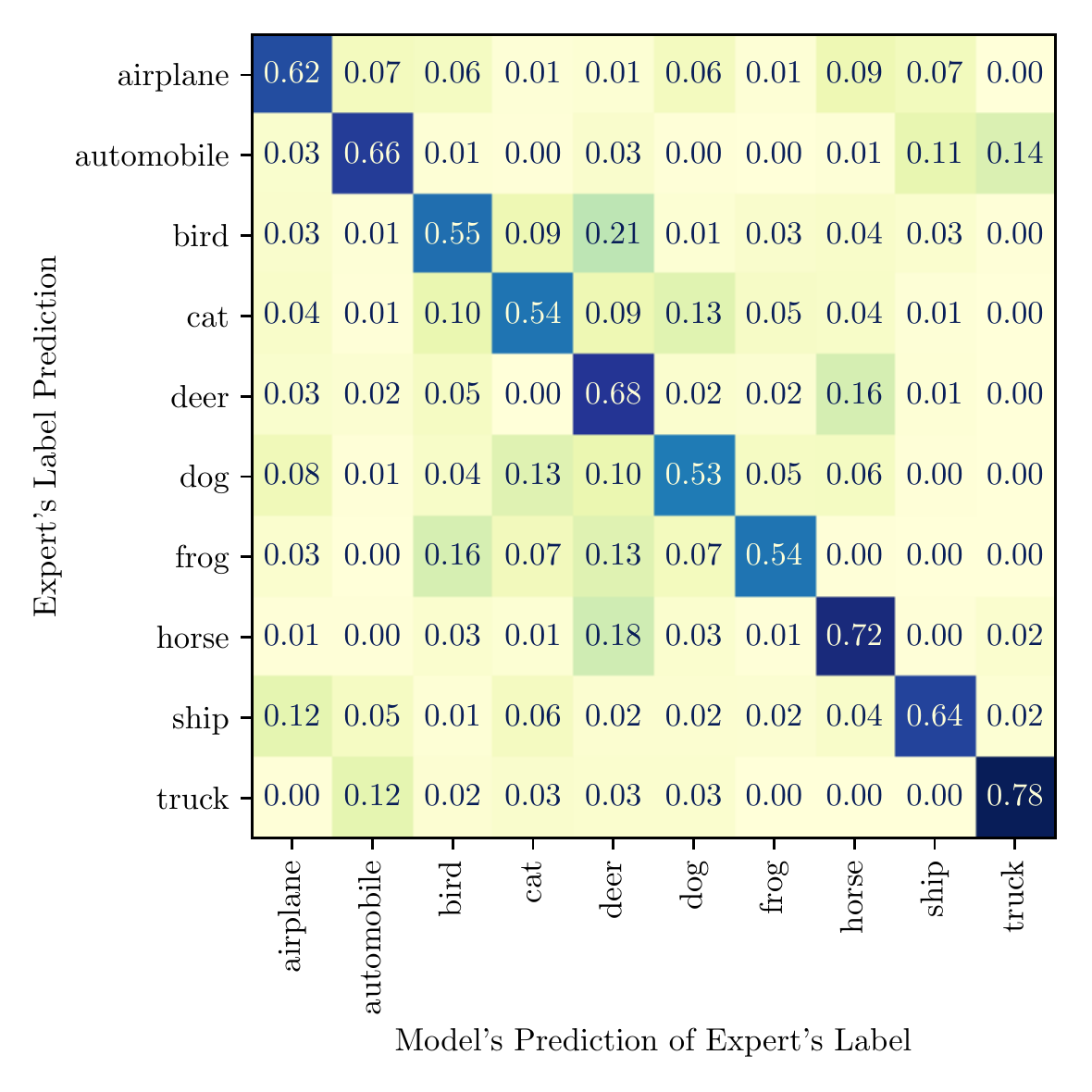}}
        \caption{
		Confusion matrices of the counterfactual predictions of our model and the predictions of the two baselines for expert's labels on the test dataset.
        }
\label{fig:confusion_matrices}
\end{figure*}

\end{document}